\pdfoutput=1
\documentclass[letterpaper, 10 pt, conference]{ieeeconf}  

\IEEEoverridecommandlockouts                              

\usepackage{cite}
\usepackage{graphicx}
\usepackage{subcaption}
\usepackage{amsmath}
\usepackage{booktabs}
\usepackage{multirow}
\usepackage{url}
\usepackage{xcolor}
\usepackage{color, soul}
\usepackage{blindtext}
\usepackage{hyperref}
\usepackage{amssymb}
\usepackage{amsfonts} 

\usepackage[flushleft]{threeparttable}
                     
\definecolor{000}{RGB}{0, 0, 0}
\definecolor{edit}{RGB}{0, 0, 255}

\newcommand{\rom}[1]{\uppercase\expandafter{\romannumeral #1\relax}}
\DeclareMathSymbol{\shortminus}{\mathbin}{AMSa}{"39}
\title{\LARGE \bf
Patchwork++: Fast and Robust Ground Segmentation \\ Solving Partial Under-Segmentation Using 3D Point Cloud
}

\author{Seungjae Lee$^{1,*}$,~\IEEEmembership{Student Member, IEEE}, Hyungtae Lim$^{1,*}$,~\IEEEmembership{Student Member, IEEE},\\ and Hyun Myung$^{1,\dagger}$,~\IEEEmembership{Senior Member, IEEE}
\thanks{*Both authors have equally contributed.}
\thanks{$^{1}$Seungjae Lee, $^{1}$Hyungtae Lim, and $^{1}$Hyun Myung are with the School of Electrical Engineering at KAIST~(Korea Advanced Institute of Science and Technology), Daejeon, 34141, Republic of Korea. {\tt\small \{sj98lee, shapelim, hmyung\}@kaist.ac.kr}}%
\thanks{$\dagger$ Corresponding author: Hyun Myung}
\thanks{This work was supported by the Industry Core Technology Development Project, 20005062, Development of Artificial Intelligence Robot Autonomous Navigation Technology for Agile Movement in Crowded Space, funded by the Ministry of Trade, Industry \& Energy (MOTIE, Republic of Korea) and BK21 FOUR.}}

\begin{document}

\captionsetup[figure]{labelformat={default},labelsep=period,name={fig.}}
\captionsetup[table]{labelformat={default},labelsep=period,name={table.}}

\maketitle
\thispagestyle{empty}
\pagestyle{empty}

\begin{abstract}


In the field of 3D perception using 3D LiDAR sensors, ground segmentation is an essential task for various purposes, such as traversable area detection and object recognition. Under these circumstances, several ground segmentation methods have been proposed. However, some limitations are still encountered. First, some ground segmentation methods require fine-tuning of parameters depending on the surroundings, which is excessively laborious and time-consuming. Moreover, even if the parameters are well adjusted, a partial under-segmentation problem can still emerge, which implies ground segmentation failures in some regions. Finally, ground segmentation methods typically fail to estimate an appropriate ground plane when the ground is above another structure, such as a retaining wall. To address these problems, we propose a robust ground segmentation method called \textit{Patchwork++}, an extension of Patchwork. Patchwork++ exploits adaptive ground likelihood estimation~(A-GLE) to calculate appropriate parameters adaptively based on the previous ground segmentation results. Moreover, temporal ground revert~(TGR) alleviates a partial under-segmentation problem by using the temporary ground property. Also, region-wise vertical plane fitting~(R-VPF) is introduced to segment the ground plane properly even if the ground is elevated with different layers. Finally, we present reflected noise removal~(RNR) to eliminate virtual noise points efficiently based on the 3D LiDAR reflection model. We demonstrate the qualitative and quantitative evaluations using a SemanticKITTI dataset. Our code is available at
\href{https://github.com/url-kaist/patchwork-plusplus}{\footnotesize\texttt{https://github.com/url-kaist/patchwork-plusplus}} 


\end{abstract}

\vspace{-0.2cm}

\section{Introduction}
\label{section:intro}


Recently, mobile robots are increasingly being employed to perform specific tasks in diverse and complex environments. Accordingly, to accomplish given tasks in various environments, several studies have improved the perception of the surrounding environments using a 3D light detection and ranging (LiDAR) sensor~\cite{lim2021patchwork,lim2021erasor}. Because most 3D LiDAR sensors are omnidirectional sensors that can accurately measure distances up to approximately 100 m, they are widely used for various purposes such as detection~\cite{li2021lidar}, segmentation~\cite{qi2017pointnet}, localization~\cite{sung2021if}, and 3D registration~\cite{koide2021voxelized}. Among the various algorithms in 3D perception via 3D LiDAR sensors, ground segmentation is one of the essential tasks for 3D perception. For example, ground segmentation can be employed in traversable area detection~\cite{suger2015traversability} and object clustering~\cite{yan2017online}.

\begin{figure}[t!]
	\centering 
	\begin{subfigure}[b]{0.23\textwidth}
		\includegraphics[width=1.0\textwidth]{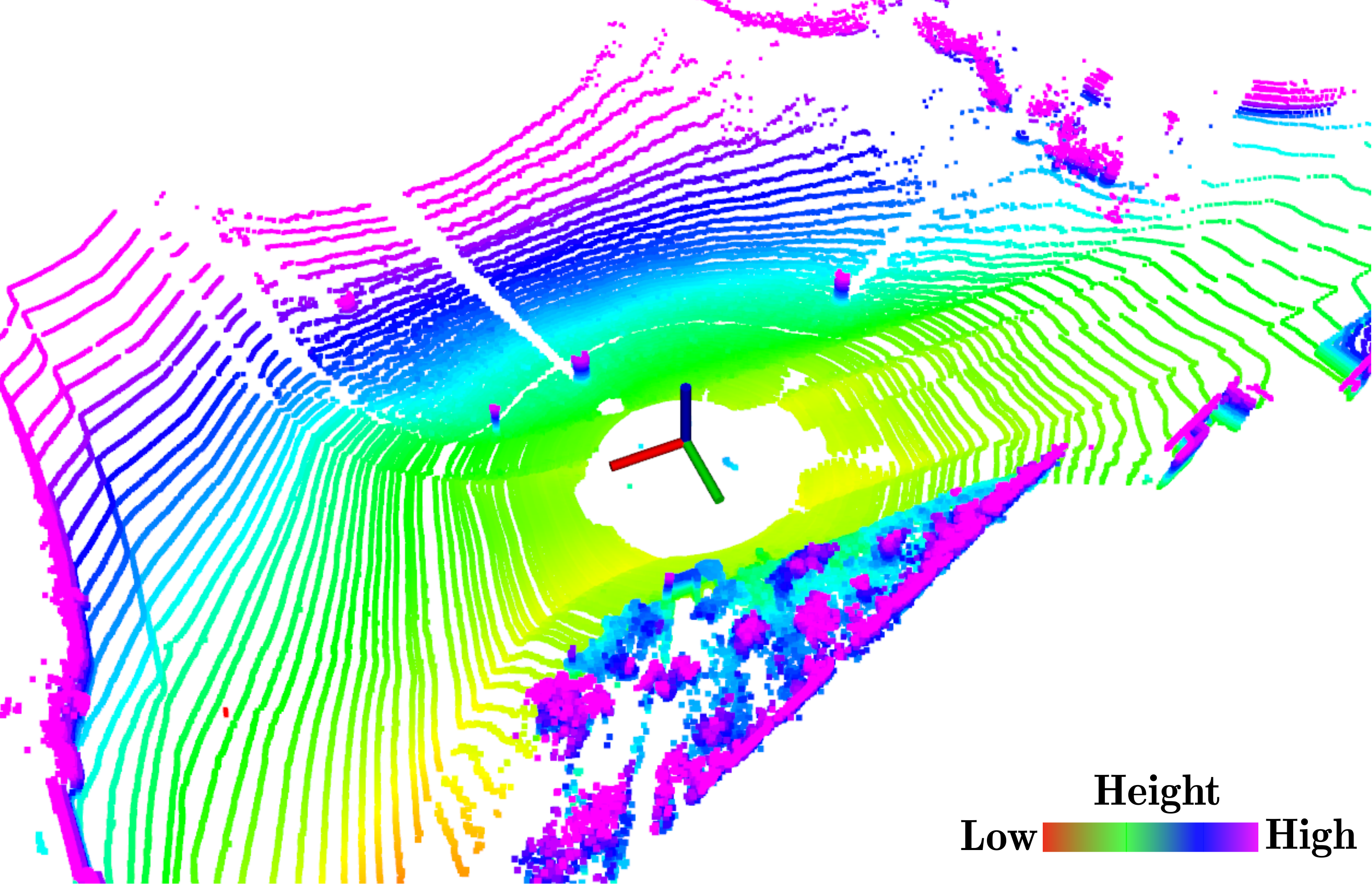}
		\caption{3D Point Cloud}
	\end{subfigure}
	\begin{subfigure}[b]{0.23\textwidth}
		\includegraphics[width=1.0\textwidth]{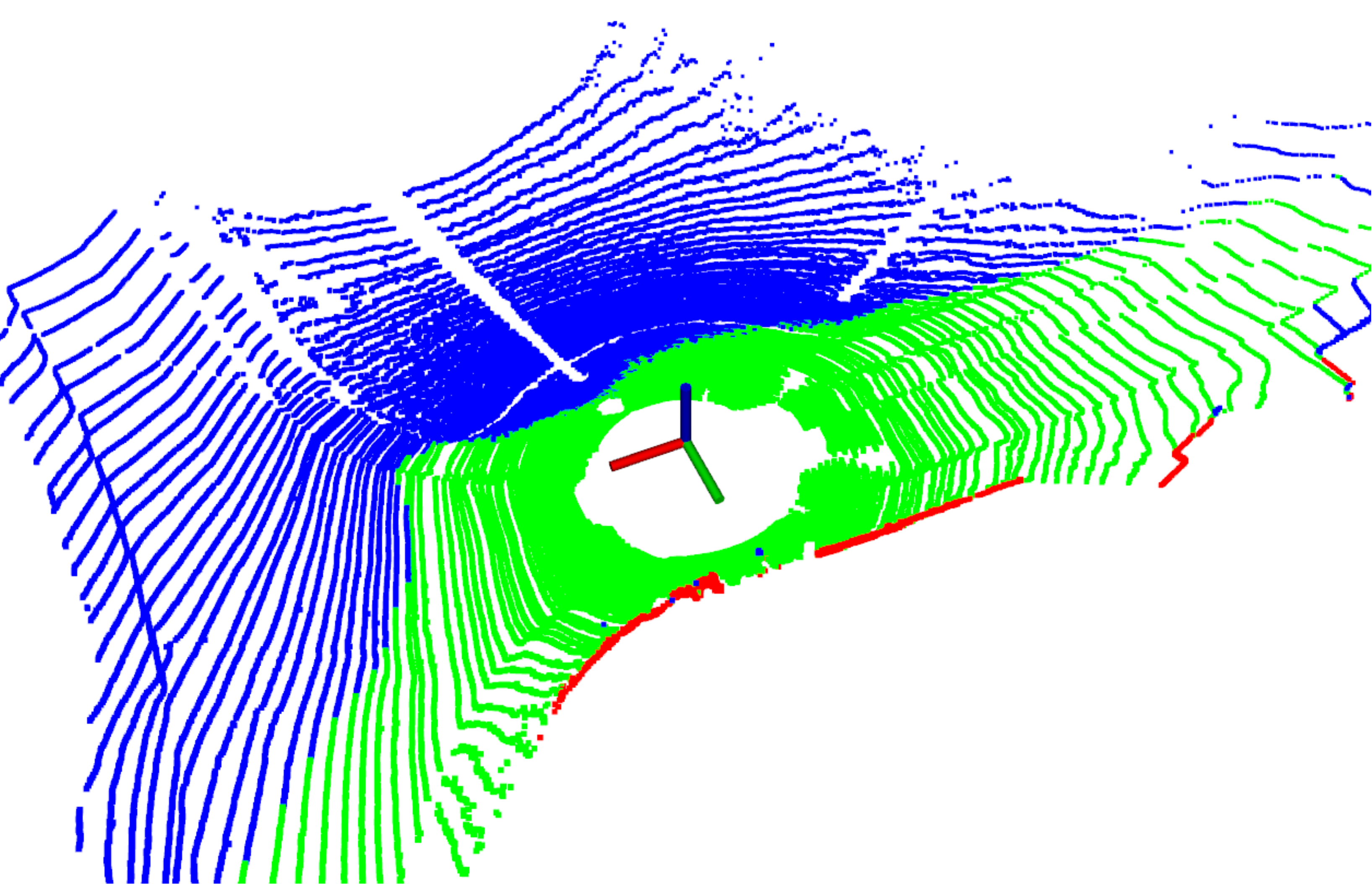}
		\caption{RANSAC~\cite{fischler1981random}}
	\end{subfigure}
	\begin{subfigure}[b]{0.23\textwidth}
		\includegraphics[width=1.0\textwidth]{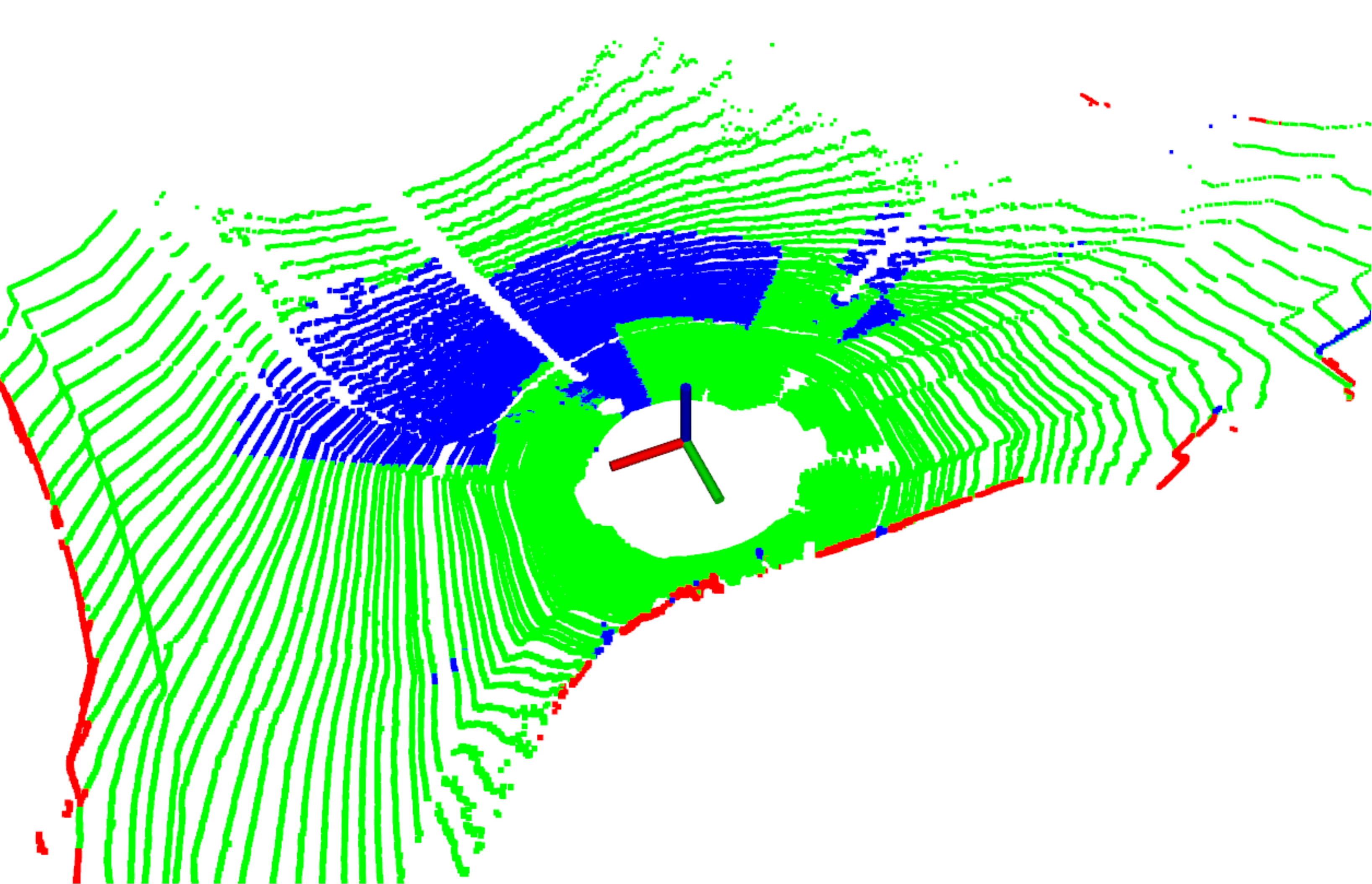}
	    \caption{Patchwork~\cite{lim2021patchwork}}
	\end{subfigure}
	\begin{subfigure}[b]{0.23\textwidth}
		\includegraphics[width=1.0\textwidth]{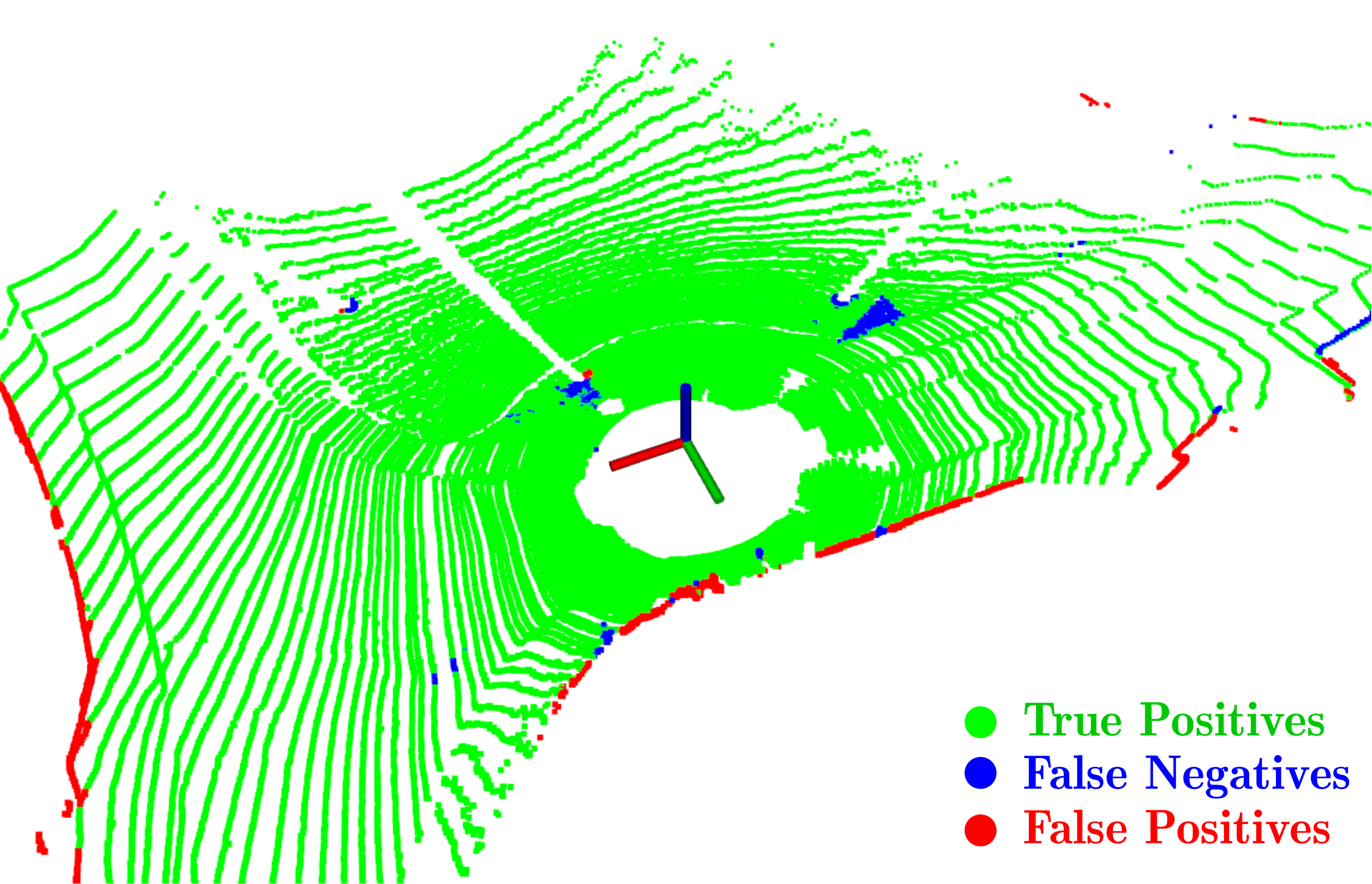}
		\caption{Patchwork++~(Ours)}
	\end{subfigure}
    \captionsetup{font=footnotesize}
	\caption{Visual descriptions of our proposed method and generic problems on ground segmentation. (a) In the urban environment, the ground may not be flat, so (b) ground segmentation using a single planar model sometimes leads to lots of false negatives called under-segmentation. (c) Patchwork~\cite{lim2021patchwork}, our previous work, handles under-segmentation, yet it sometimes fails partially. This is because its parameters are deterministic; hence, these user-defined parameters may filter out the solid ground. (d) Unlike Patchwork, Patchwork++ updates its parameters based on previous ground segmentation; therefore, Patchwork++ successfully performs ground segmentation in uneven surroundings. The red, green, and blue axes represent $x$, $y$, and $z$ coordinates, respectively.}
	\label{undersegmentation}
    \vspace{-0.5cm}
\end{figure}

As machine learning has been applied to diverse research areas, various learning-based 3D perception methods have been proposed, exhibiting outstanding performance~\cite{xu2021rpvnet,paigwar2020gndnet}. But, unfortunately, there are two inherent potential limitations. First, learning-based methods require point-wise labeling by humans, which is both time-consuming and severely laborious. Second, the performance of learning-based ground segmentation can potentially be degraded when employed in scenes different from the training dataset or different sensor configurations. Consequently, in this study, we focus on ground segmentation in non-learning-based methods.


In general, ground segmentation functions as a preceding stage because the ground and non-ground points can be utilized for different purposes. For example, non-ground points go through another segmentation stage, which classifies these non-ground points into static and dynamic points. However, because ground points can be assumed as being static, the corresponding segmentation is not performed on the ground points.

Moreover, when ground segmentation is exploited as a preprocessing step, it should satisfy the following three requirements. First, ground segmentation should be sufficiently fast to be employed as a preliminary process for other algorithms such as object clustering. Second, it should ensure little performance perturbation to simultaneously guarantee both precision and recall performance. 
Third, it should be capable of dealing with uneven outdoor environments.



However, several ground segmentation methods fail to satisfy these requirements and even exhibit the under-segmentation problem, as represented in Fig.~\ref{undersegmentation}(b). Although our previous work, Patchwork~\cite{lim2021patchwork}, meets these requirements, it occasionally exhibits a \textit{partial under-segmentation problem}, as represented in Fig.~\ref{undersegmentation}(c). These under-segmentation issues usually occur because of two potential reasons. First, some ground segmentation methods are deterministic.
Second, there are some situations in which the assumption that the points with low $z$ values are most likely to be considered as the ground points does not hold.

Therefore, \textit{Patchwork++}, an advanced version of Patchwork, is proposed to tackle these problems. Accordingly, our proposed method becomes more generalized in complex urban environments. Our contributions are threefold as follows:

\begin{itemize}
    \item First, adaptive ground likelihood estimation~(A-GLE) and temporal ground revert~(TGR) are proposed, which provide even fewer false negatives than Patchwork based on the \textcolor{000}{coarse}-to-fine strategy that addresses the partial under-segmentation issue.
    \item Second, two novel outlier rejection modules, i.e., reflected noise removal~(RNR) and region-wise vertical plane fitting~(R-VPF), are proposed to prevent the situations where our assumption does not hold by rejecting noises or non-ground points.
    \item Third, our proposed algorithm was evaluated with the SemanticKITTI dataset \cite{behley2019semantickitti} and experimental evidence corroborates Patchwork++ exhibits promising performance compared with the state-of-the-art methods including Patchwork and shows the fastest speed as well.
\end{itemize}

\vspace{-0.15cm}
\section{Related Works}
\subsection{Learning-based Ground Segmentation Methods}
Xu \textit{et al.} \cite{xu2021rpvnet} proposed a novel range-point-voxel fusion network, RPVNet, which ranks first on the SemanticKITTI leaderboard of semantic segmentation. RPVNet appropriately combines three different features from range image, point, and voxel to classify each point. Shen \textit{et al.} \cite{shen2021fast} addressed the problem of the capability of segmentation on rough terrains and the efficiency of processing time by introducing the Jump-Convolution-Process, which facilitates the point cloud segmentation by solving the 2D smoothing problem. Paigwar \textit{et al.} \cite{paigwar2020gndnet} proposed GndNet, which estimates the grid-wise ground elevation and segments ground and non-ground points using PointNet \cite{qi2017pointnet} to encode a pillar feature in each grid. As mentioned in Section~\ref{section:intro}, there are several potential limitations in that these methods require human-made labeling by hands and the performance can be degraded when it comes to unforeseen situations that are not included in the training data.

\subsection{Conventional Ground Segmentation Methods}

Moosmann \textit{et al.} \cite{moosmann2009segmentation} proposed a 3D point cloud segmentation method based on the idea of \textit{local convexity} yielding a fast ground segmentation in a non-flat urban environment. Himmelsbach \textit{et al.} \cite{himmelsbach2010fast} also proposed a line-based ground segmentation method providing a \textcolor{000}{significantly} high speed. Zermas \textit{et al.} \cite{zermas2017fast} proposed a fast ground plane fitting~(GPF), which utilizes an initial seed extraction suitable for the ground plane estimation based on principal component analysis~(PCA). Unlike other ground segmentation methods, such as RANSAC~\cite{fischler1981random}, this approach circumvents redundant random iterations, which are severely time-consuming. 
As an extension of GPF, Lim \textit{et al.} \cite{lim2021patchwork} proposed a concentric zone-based region-wise ground segmentation with the GLE. Narksri \textit{et al.} \cite{narksri2018slope} \textcolor{000}{proposed} a RANSAC-based ground segmentation with \textcolor{000}{multiple regions} and addressed the problem posed by sloped terrain. Jim\'{e}nez \textit{et al.} \cite{jimenez2021ground} addressed the issue of local geometric variation of the ground by adopting the channel-based Markov random field.


Although various conventional ground segmentation methods have been proposed, severe problems remain inherent in these methods. In particular, under-segmentation is the representative problem that results in lots of false negatives, as represented in Fig.~\ref{undersegmentation}. For example, single planar model-based methods usually fail to estimate the ground plane with a slope because the single plane assumption is insufficient to model wide areas with partially bumpy regions as a single plane. Although Narksri \textit{et al.} \cite{narksri2018slope} addressed this problem by employing a RANSAC-based ground plane fitting in multiple regions, it still exhibits the under-segmentation problem because of its limitation in dealing with bumpy grounds.




\subsection{Applications of Ground Segmentation}

As mentioned in Section~\ref{section:intro}, various object clustering methods \cite{li2020insclustering,sualeh2019dynamic, asvadi20163d,jimenez2021ground} include ground segmentation as a prior step for object recognition. 
Because ground points are not the region of interest in object clustering methods, the elimination of ground points enhances both computational efficiency and accuracy. That is, approximately half of the points in a 3D LiDAR scan are given as ground points in outdoor environments. This implies that the overall algorithm can be faster, as the size of the input point cloud will be half of the original via ground removal.

In addition, ground segmentation is also employed in LiDAR odometry methods to obtain strong constraints using ground points as features~\cite{shan2018lego,pan2021mulls,lim2022pago}. For instance, Shan \textit{et al.} \cite{shan2018lego} proposed LeGO-LOAM, a lightweight and \textcolor{000}{ground-optimized} LiDAR SLAM. According to the research, a few planar features are extracted from the ground to estimate the transformation in terms of \textit{z}~(height direction), roll angle, and pitch angle. Furthermore, Pan \textit{et al.} \cite{pan2021mulls} proposed MULLS, a versatile LiDAR SLAM method using the multi-metric linear least square. That is, MULLS separates points into several classes based on their geometric features to construct the multi-metric optimization problem. Certainly, the ground is also considered as one of the several types.

\section{Patchwork++: Fast, Robust, and Adaptive Ground Segmentation}

\begin{figure}[t]
	\centering
    \vspace{0.15cm} 
	\includegraphics[width=0.46\textwidth]{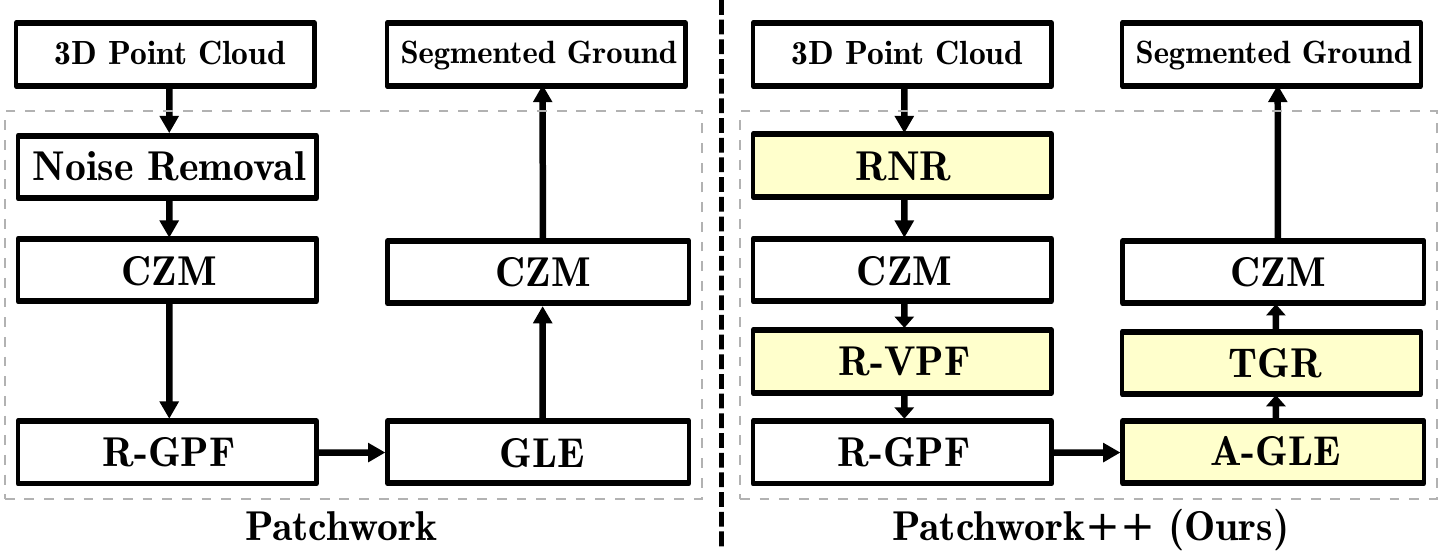}
    \captionsetup{font=footnotesize}
    \vspace{-0.1cm}
	\caption{(L-R) Flow charts of Patchwork~\cite{lim2021patchwork} and Patchwork++. Unlike Patchwork, Patchwork++ consists of novel modules called Reflected Noise Removal~(RNR), Region-wise Vertical Plane Fitting~(R-VPF), Adaptive GLE~(A-GLE), and Temporal Ground Revert~(TGR). CZM and R-GPF represent the concentric zone model and region-wise ground plane fitting, respectively, which were proposed in the previous work \cite{lim2021patchwork}.}
	\label{fig:flow_chart}
	\vspace{-0.6cm}
\end{figure}

The following paragraphs highlight the problem definition and rationale behind each module of Patchwork++. Unlike Patchwork, the following four parts are proposed: RNR, R-VPF, A-GLE, and TGR, which are presented in Fig.~\ref{fig:flow_chart}.

\subsection{Problem Definition}

Given a 3D point cloud, $P$, which represents the set of measured points, $p_i$, and presents a snapshot of the surroundings, we attempt to classify every $p_i$ into the ground, including road, terrain, and sidewalks, or non-ground class, which includes cars, buildings, humans, etc. 
Then, $P$ can be represented as a union of two distinct sets: a set of ground points, $G$, and non-ground points, $N$. In contrast, a ground segmentation method outputs the estimated ground points, $\hat{G}$, while the remaining points are regarded as a set of non-ground points, $\hat{N}$. Therefore, $P$ can be expressed as follows:
\begin{equation}
    P = \bigcup\limits_{i}{p_i} = {G\cup N} = {\hat{G}\cup \hat{N}}.
\end{equation}

In this case, $\hat{G}$ and $\hat{N}$ can be divided into actual ground points and actual non-ground points, respectively. Specifically, ${\hat{G}}\cap{G}$, ${\hat{G}}\cap{N}$, ${\hat{N}}\cap{G}$, and ${\hat{N}}\cap{N}$ represent the set of true positives~($TP$), false positives~($FP$), false negatives~($FN$), and true negatives~($TN$), respectively.
In conclusion, the objective of this study is to estimate $\hat{G}$ that contains $TP$ points as many as possible while rejecting $FN$ and $FP$ points as many as possible.

\subsection{RNR: Reflected Noise Removal}
\label{section:rnr}

\begin{figure}[b!]
    \captionsetup{font=footnotesize}
	\centering 
	\vspace{-0.3cm}
	\includegraphics[width=0.48\textwidth]{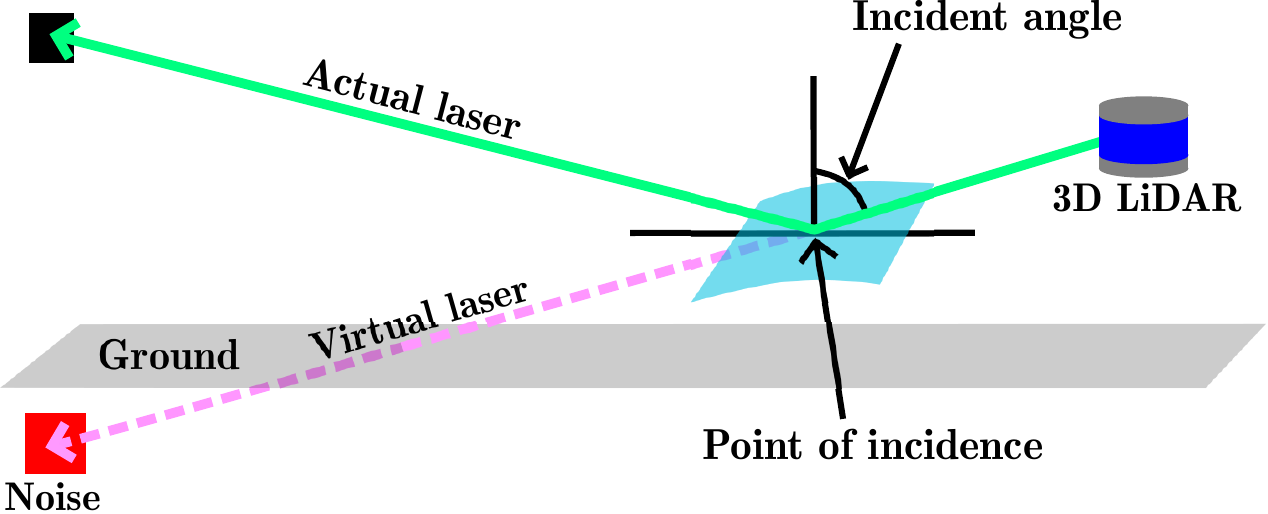}
	\vspace{-0.5cm}
	\caption{Illustration of reflected noises in a 3D point cloud. Noise~(red square) can be located below the ground due to the reflection of bottom lasers.}
	\label{fig:ray}
\end{figure}

For a given raw point cloud, some noise points are occasionally observed under the actual ground, as illustrated in Fig.~\ref{fig:ray}.
Unfortunately, R-GPF selects initial seeds based on the assumption that the points at the lowest part in the bin are ground~\cite{zermas2017fast}; hence, these noises can be regarded as initial seeds that trigger the under-segmentation problem in the corresponding bin.

\begin{figure}[t!]
    \vspace{0.15cm}
	\centering 
	\begin{subfigure}[b]{0.23\textwidth}
		\includegraphics[width=1.0\textwidth]{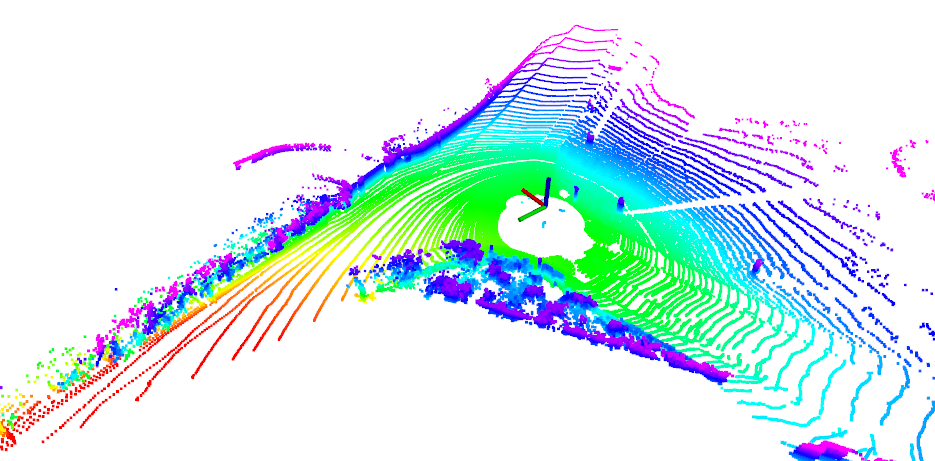}
		\caption{}
		\label{downhill_a}
	\end{subfigure}
	\begin{subfigure}[b]{0.23\textwidth}
		\includegraphics[width=1.0\textwidth]{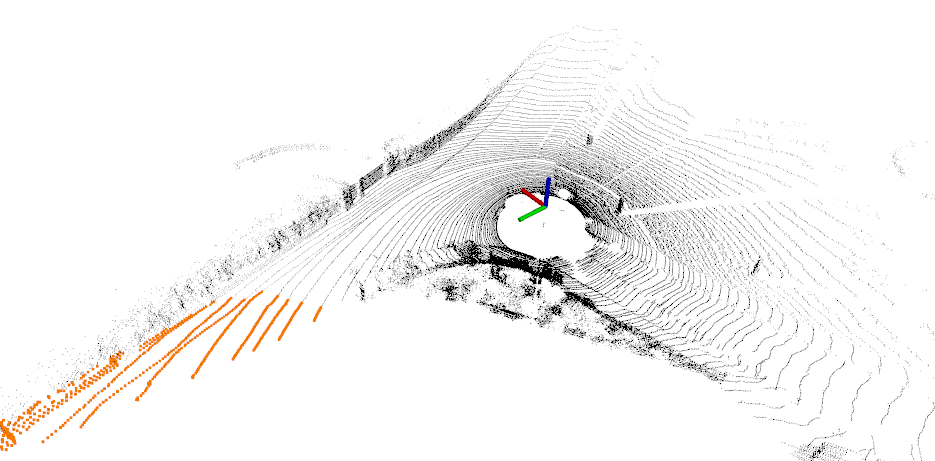}
		\caption{}
		\label{downhill_b}
	\end{subfigure}
	\begin{subfigure}[b]{0.23\textwidth}
		\includegraphics[width=1.0\textwidth]{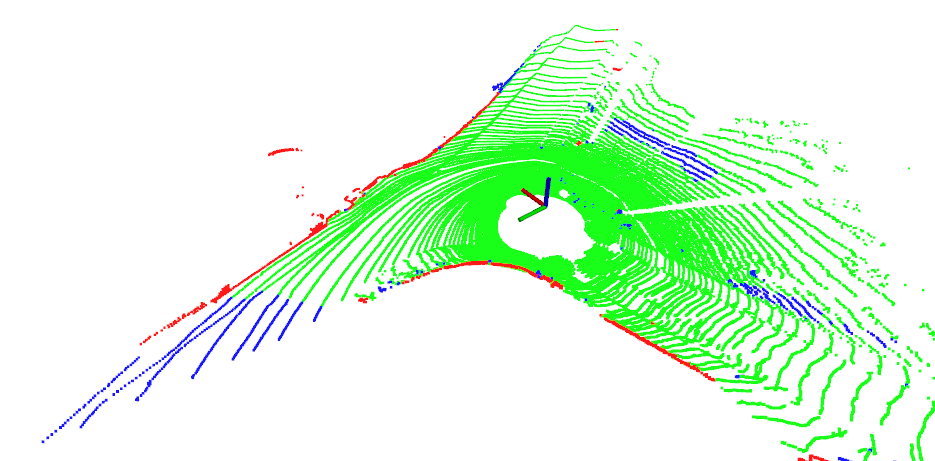}
		\caption{}
		\label{downhill_c}
	\end{subfigure}
	\begin{subfigure}[b]{0.23\textwidth}
		\includegraphics[width=1.0\textwidth]{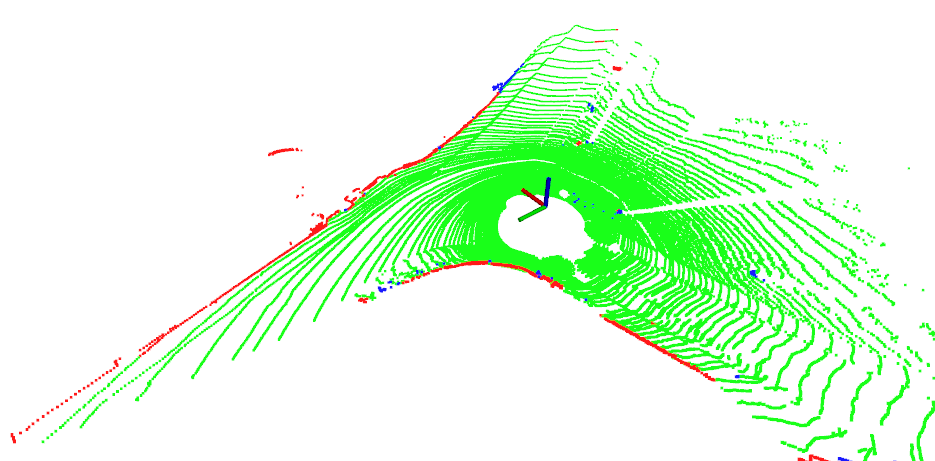}
		\caption{}
		\label{downhill_d}
	\end{subfigure}
    \captionsetup{font=footnotesize}
    \vspace{-0.1cm}
	\caption{(a) A 3D point cloud visualized by an elevation map, where red points represent the bottom part of the downhill. (b) Height threshold-based noise filtering triggers several false negatives because $z$ values of some ground points are lower than $z_\text{min}$ (orange parts). (c)–(d) Before and after the application of our RNR: (c)~In Patchwork~\cite{lim2021patchwork}, some ground points are under-segmented (blue), (d)~whereas our proposed method successfully identifies the points from downhill as ground points (best viewed in color). }
	\label{downhill}
	\vspace{-0.6cm}
\end{figure}

To address this problem, these points are usually filtered out by rejecting points whose $z$ value is smaller than a user-defined threshold, $z_{\text{min}}$~\cite{lim2021erasor,lim2021patchwork}. This filtering assumes that the ground of surroundings is sufficiently flat. Unfortunately, this noise removal method also triggers a severe problem that erases a few true ground points undesirably in steep slope environments, as illustrated in Figs.~\ref{downhill}(b) and~(c). Therefore, it is necessary to discriminate between the actual ground points and noise points.




To this end, RNR is proposed by leveraging two key observations \textcolor{000}{based} on the research by Zhao \textit{et al.}~\cite{zhao2020mapping}. First, the virtual noise points can be obtained by the reflection on the reflective surfaces, i.e., bonnets and roofs of vehicles, or glass. This fact lets the noise points be located along the direction from the sensor origin onto the point of incidence, as shown in Fig.~\ref{fig:ray}.
It is noticeable that the smaller the incident angle, the lower the $z$ value of the noise tends to occur. For this reason, it can be concluded that critical noises that hinder our assumption are highly likely to be originated from the rays from the bottom rings because the ray directions of the bottom rings tend to have relatively small incident angles.

Second, these noise points have relatively small intensities in general because the rays which correspond to the noise points undergo more reflections than the usual rays until they return to the sensor. 
For example, the ray of the virtual point takes additional reflection on the point of incidence, as illustrated in Fig.~\ref{fig:ray}. Consequently, this additional reflection triggers the decrease of the intensity of the virtual point.

Therefore, firstly, RNR only checks a certain number of bottom rings, $N_{\text{noise}}$, to remove these noise points. Accordingly, RNR rejects the points that are located in these rings with a lower height than the noise removal height threshold, $h_{\text{noise}}$, and lower intensity than the noise removal intensity threshold, $I_{\text{noise}}$. Consequently, RNR efficiently removes noise points with unrealistically large negative $z$ values with a negligible loss of actual ground points. In particular, $h_{\text{noise}}$ is self-updated by A-GLE (see Section~\rom{3}.\textit{D}).

\subsection{R-VPF: Region-wise Vertical Plane Fitting}
\label{section:r_vpf}

\begin{figure}[b!]
	\centering 
	\includegraphics[width=0.48\textwidth]{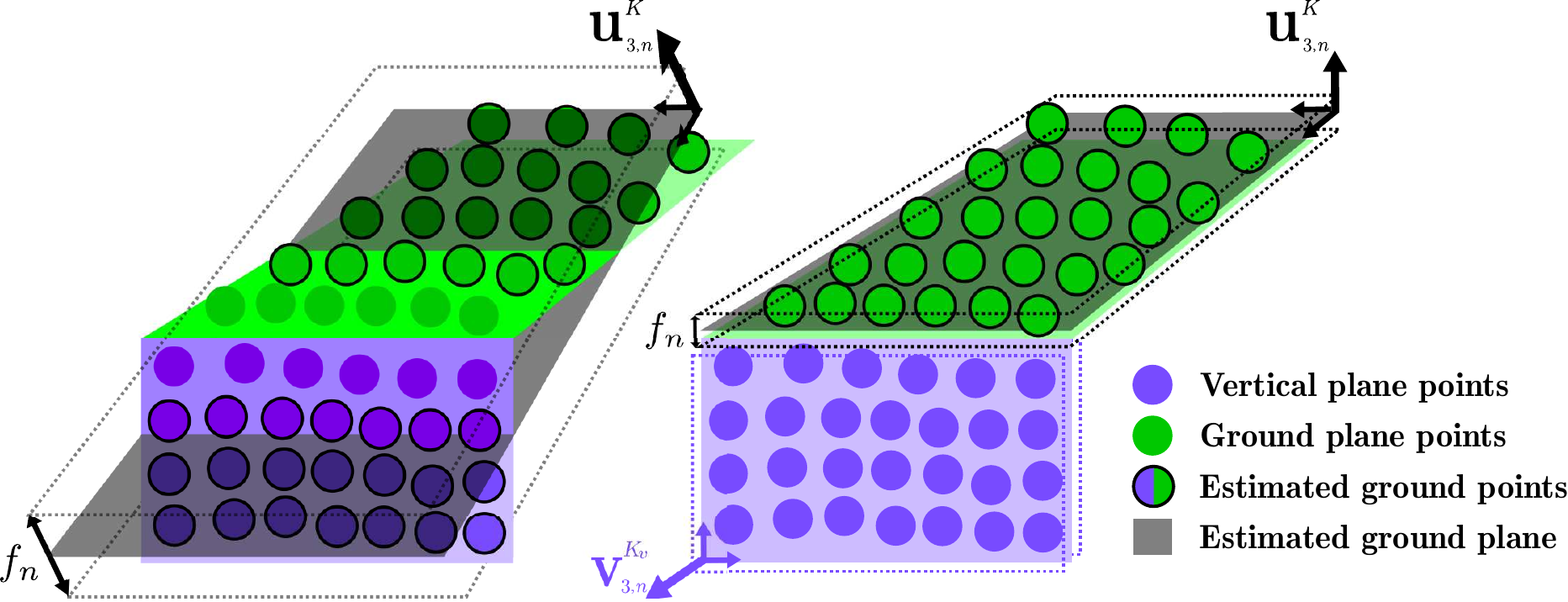}
	\captionsetup{font=footnotesize}
	\caption{(L-R): Visual descriptions of the estimated ground plane without R-VPF and with R-VPF. Once R-VPF is applied for preprocessing, R-GPF successfully estimates an accurate ground plane even though true ground points are located on the non-ground points. In addition, $f_n$, which is the flatness variable~\cite{lim2021patchwork} of the estimated plane, becomes smaller. Accordingly, the estimated ground can be distinguishable as a ground owing to its small~$f_n$.}
	\label{fig:rvpf_illust}
\end{figure}

However, R-GPF still sometimes fails in estimating an appropriate ground plane in another situation where the desired ground is above vertical urban structures. This is because the points from the vertical structures are chosen as the initial seeds due to their smaller $z$ values compared with actual ground points. 
For example, when retaining walls are located under the ground in the bin and close enough to the sensor, R-GPF selects the points from the wall as the initial seeds of the ground. Then, these points induce an unsuitable ground fitting result with the ground plane points on them because PCA-based estimation is significantly sensitive to the outlier points, as illustrated in Fig.~\ref{fig:rvpf_illust}. 

One may argue that this elevated ground should be considered as non-ground because this ground is distinguishable from another ground and the points from this ground have larger $z$ values than points from \textcolor{000}{the} usual ground. However, some dynamic objects, such as humans, can still stand on that ground; thus, these points are considered actual ground points rather than non-ground points in this study.




To this end, R-VPF is proposed\textcolor{000}{, allowing} R-GPF to estimate more accurate ground planes by rejecting dominant and vertical points for preprocessing within a bin. Specifically, let $P_n$ be the points in the $n$-th bin, which is divided based on the concentric zone model~(CZM)~\cite{lim2021patchwork}. That is, CZM \textcolor{000}{divides} a 3D point cloud into multiple subsets in the polar coordinate and bin sizes are different depending on which zone the bins belong to. 
Then, the estimated vertical plane points, $\hat{V}_{n}$, are determined via the following four steps. 

First, R-VPF selects some points starting from the lowest point, as seed points in every $k$-th iteration. Second, R-VPF estimates the mean of seed points, $\textbf{m}_n^k \in \mathbb{R}^3$, and the unit normal vector, $\textbf{v}_{3,n}^k$, which corresponds to the eigenvector with the smallest eigenvalue from the PCA of seed points. Subsequently, the $k$-th potential vertical plane points, $\hat{W}_{n}^k$, are sampled among the candidate points, $\hat{P}^{k}_n$, as follows:

\begin{equation}
    \hat{W}_{n}^k = \Big\{ \textbf{p} \in \hat{P}^{k}_n \ \Big| \ { |(\textbf{p}-\textbf{m}_n^k)\cdot\textbf{v}_{3,n}^k}| < d_v\Big\} ,
\end{equation}

\noindent where $\hat{P}^{k}_n \leftarrow P_n$ when $k=1$, $\hat{P}^{k}_n \leftarrow {P_n-\bigcup_{i=1}^{k-1}\hat{V}_{n}^i}$, otherwise, and $d_v$ denotes the distance margin for the estimated vertical plane. This implies that the points considered as vertical points, $\hat{V}_n^i$, in previous iterations are rejected in subsequent iterations.

Then, $\hat{V}_{n}^k$, which denotes the vertical points at the $k$-th iteration, is determined based on the following condition:


\begin{equation}
\begin{aligned}
    \hat{V}_{n}^k = \Big\{ \begin{array}{ll}
    \hat{W}_{n}^k,  &  \text{if} \; {\frac{\pi}{2}}-\cos^{-1}{(\textbf{v}_{3,n}^{k} \cdot \textbf{u}_z)} < \theta_v \\
    \varnothing,     &  \text{otherwise} 
    \end{array} ,
\end{aligned}
\end{equation}


 \noindent where $\textbf{u}_z$ denotes the unit normal vector in the $z$-axis direction, i.e., $[0 \; 0 \; 1]^\intercal$ and $\theta_v$ is the margin for the unit normal vector of the estimated plane.
 

Finally, the total vertical points, $\hat{V}_{n}$, are estimated by accumulating all the vertical points as follows:
\begin{equation}
    \hat{V}_{n} = \bigcup_{k=1}^{K_v} {\hat{V}_{n}^k} ,
\end{equation} where $K_v$ denotes the number of iterations for R-VPF.



\subsection{A-GLE: Adaptive Ground Likelihood Estimation} 
\label{section:adaptive_gle}

In our previous work~\cite{lim2021patchwork}, GLE, $f(\mathcal{X}_n | \theta_n)$, was employed to determine whether the region-wise estimated ground planes are ground or non-ground. For this process, GLE is determined by \textit{uprightness}, \textit{elevation}, and \textit{flatness} indicator functions with $\textbf{v}_{3,n}$, $\bar{z}_n$, and $\sigma_n$ as variables, respectively. In detail, $\textbf{v}_{3,n}$ is the eigenvector with the smallest eigenvalue from the PCA of seed points, $\bar{z}_n$ is the mean of seed points, and $\sigma_n$ is the local surface variation~\cite{weinmann2015semantic}, $\frac{\lambda_{3,n}}{\lambda_{1,n}+\lambda_{2,n}+\lambda_{3,n}}$, where $\lambda_{1,n}$, $\lambda_{2,n}$, and $\lambda_{3,n}$ are the eigenvalues from the PCA of given points in the descending order. 

\begin{figure}[t!]
	\centering
    \vspace{0.15cm} 
	\begin{subfigure}[b]{0.23\textwidth}
		\includegraphics[width=1.0\textwidth]{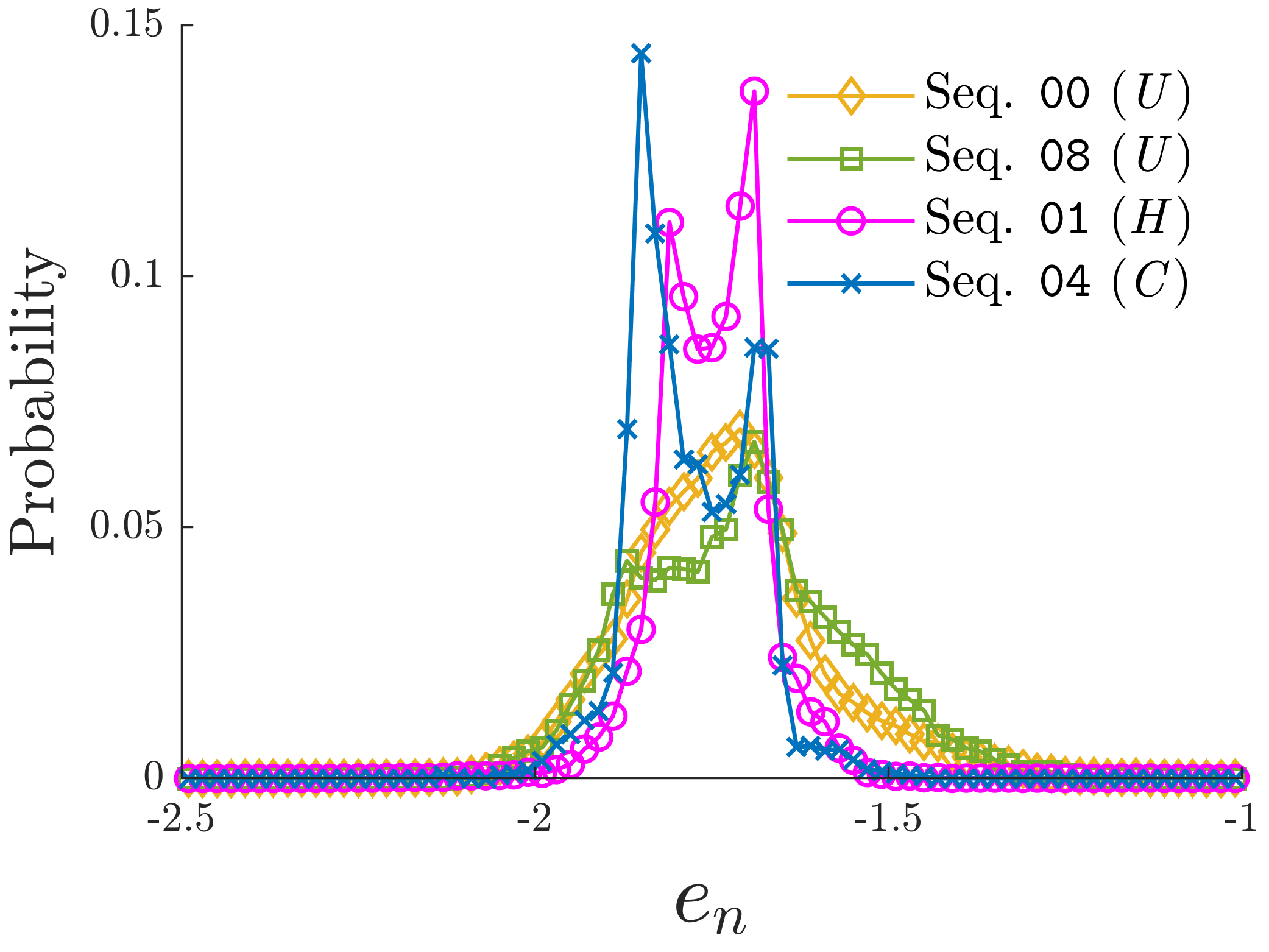}
		\caption{}
	\end{subfigure}
	\begin{subfigure}[b]{0.23\textwidth}
		\includegraphics[width=1.0\textwidth]{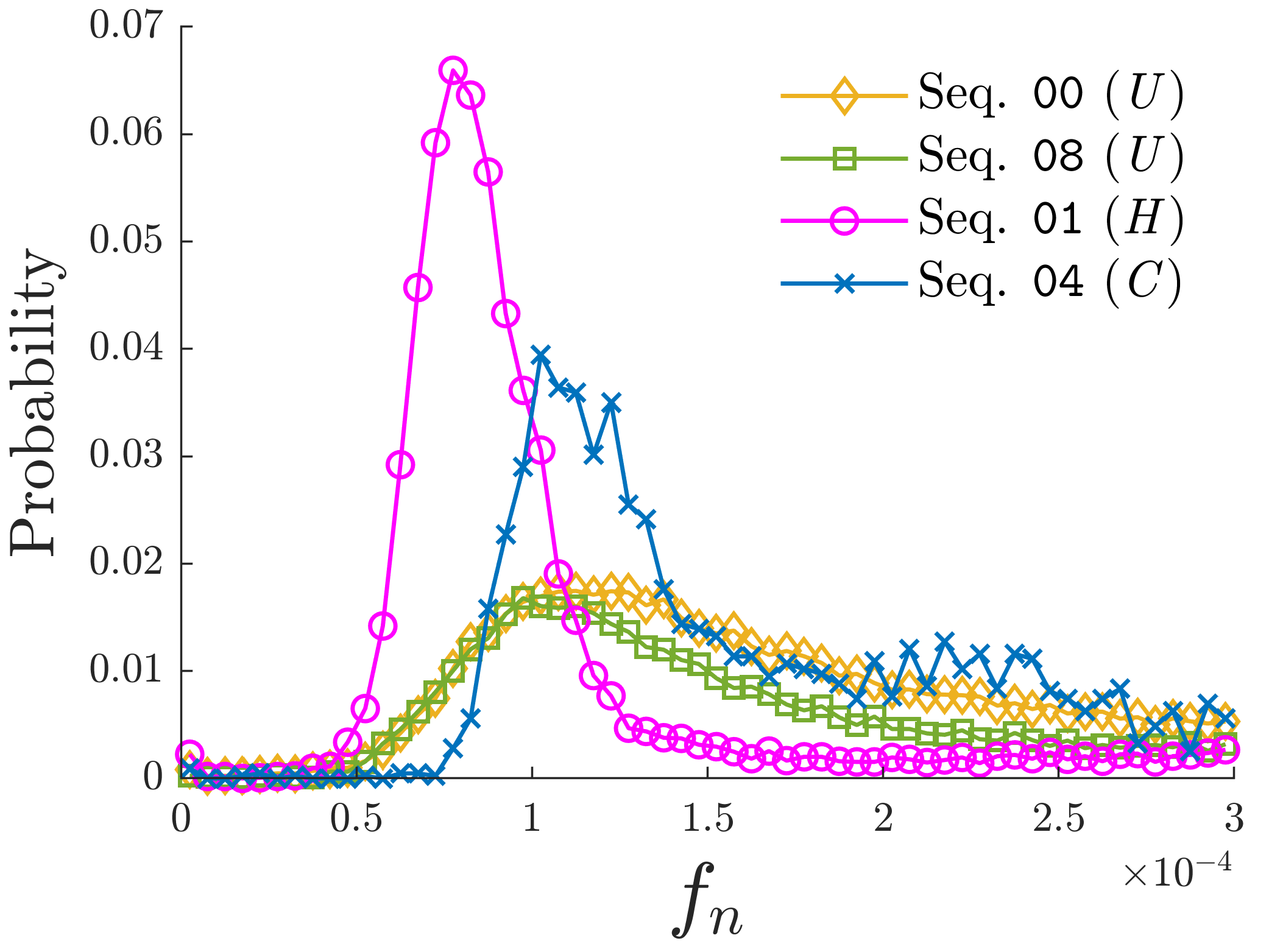}
		\caption{}
	\end{subfigure}
    \captionsetup{font=footnotesize}
    \vspace{-0.1cm}
	\caption{Probability distributions of true ground points' (a) $e_n$ and (b) $f_n$ at the first ring of the concentric zone model in Sequence \texttt{00}, \texttt{01}, \texttt{04}, and \texttt{08}. Sequence \texttt{00} and \texttt{08} represent similar distributions. In contrast, Sequence \texttt{01} and \texttt{04} result in quite different distributions from each other. \textit{U}, \textit{H}, and \textit{C} 
	denote urban, highway, and country scenes, respectively.}
	\label{fig:GT_analysis}
	\vspace{-0.5cm}
\end{figure}

Accordingly, it is necessary to set the corresponding parameters~($\theta_\tau, \kappa(r_n), \sigma_{\tau, m}$) as the thresholds of uprightness, elevation, and flatness indicator functions, respectively. In this paper, $\bar{z}_n$, $\sigma_n$, $\kappa(r_n)$\textcolor{000}{,} and $\sigma_{\tau,m}$ are represented as $e_n$, $f_n$, $e_{\tau,m}$\textcolor{000}{,} and $f_{\tau,m}$ for simplicity. However, in our previous work, setting the optimal parameters, especially $e_{\tau,m}$ and $f_{\tau,m}$, depending on the surroundings is severely time-consuming. For example, in Fig.~\ref{fig:GT_analysis}, the distributions of elevation and flatness values for true ground planes seem to be varied depending on the surroundings. Hence, this confirms that the adjustment of parameters is necessary for each environment. 

To tackle this problem, A-GLE updates $e_{\tau,m}$ and $f_{\tau,m}$ in an adaptive manner based on the previously obtained estimation results. Unless A-GLE is initialized with abnormal values of $e_{\tau,m}$ and $f_{\tau,m}$, A-GLE can update these parameters properly. In addition, as mentioned in Section~\rom{3}.\textit{\textcolor{000}{B}}, the noise removal height threshold, $h_{\text{noise}}$, is updated by A-GLE.

\noindent \textbf{Elevation} \; For the self-update of parameters, we propose a novel concept of \textit{definite ground}, $D_m$, which represents the previously estimated ground planes satisfying uprightness and elevation conditions in the $m$-th ring of the concentric zone model. Accordingly, A-GLE adopts the properties of definite ground planes to update the elevation parameter for the next estimation.

Let $E_m$ be the set of all $e_n$ values of estimated ground planes in $D_m$. Then, based on $E_m$, $e_{\tau,m}$ is newly updated as follows:
\begin{equation}
    e_{\tau,m} \leftarrow \text{mean} \big( E_{m} \big) + a_m\cdot \text{stdev}\big( E_{m} \big) ,
    \label{eqn:self_update_thr_e}
\end{equation} where $\text{mean}(\cdot)$ and $\text{stdev}(\cdot)$ denote the mean and standard deviation of a given set, respectively. $a_m>0$ denotes the constant gain of the standard deviation term in the $m$-th ring of the concentric zone model for elevation.

\noindent \textbf{Flatness} \; Before updating the flatness parameter $f_{\tau,m}$, it is found that a potential problem exists in its calculation in our previous work. Specifically, as mentioned above, $f_n$ was calculated as $\frac{\lambda_{3,n}}{\lambda_{1,n}+\lambda_{2,n}+\lambda_{3,n}}$. To obtain a meaningful value from this approach, the value must be consistent for the same ground regardless of the ground's relative LiDAR position.

Unfortunately, because Patchwork leverages the concentric zone model, the ground is divided into several bins with the sensor origin as the center. Accordingly, the shape and size of the ground can be differentiated by the approach taken to divide the ground. Then, $\lambda_{1,n}$ and $\lambda_{2,n}$ can be changed even though the overall ground does not change. Therefore, one can argue that $f_n$ calculated accordingly is illogical based on its potential inconsistency. To address this problem, Patchwork++ sets $f_n=\lambda_{3,n}$, which is a variation in the surface normal direction.



Similar to $e_{\tau,m}$, $f_{\tau,m}$ has been updated based on the distribution of $D_m$ by A-GLE. Let $F_m$ be the set of all $f_n$ for the estimated ground planes in $D_m$. Then, $f_{\tau,m}$ can be updated as follows:
\begin{equation}
    f_{\tau, m} \leftarrow \text{mean} \big( F_{m} \big) + b_m\cdot \text{stdev} \big( F_{m} \big) ,
    \label{eqn:self_update_thr_f}
\end{equation} where $b_m>0$ denotes the constant gain of the standard deviation term in the $m$-th ring of the concentric zone model for flatness. 

\noindent \textcolor{000}{\textbf{Noise Removal Height} \; A-GLE updates $h_{\text{noise}}$ based on the mean of the elevation values of the estimated ground planes at the closest ring of the concentric zone model as follows:
\begin{equation}
    h_{\text{noise}} \leftarrow \text{mean} \big( E_1 \big) + \delta
    \label{eqn:noise_height} ,
\end{equation} where $\delta<0$ means the distance margin of the noise removal.}

\begin{figure}[b!]
    \captionsetup{font=footnotesize}
	\centering 
	\includegraphics[width=0.46\textwidth]{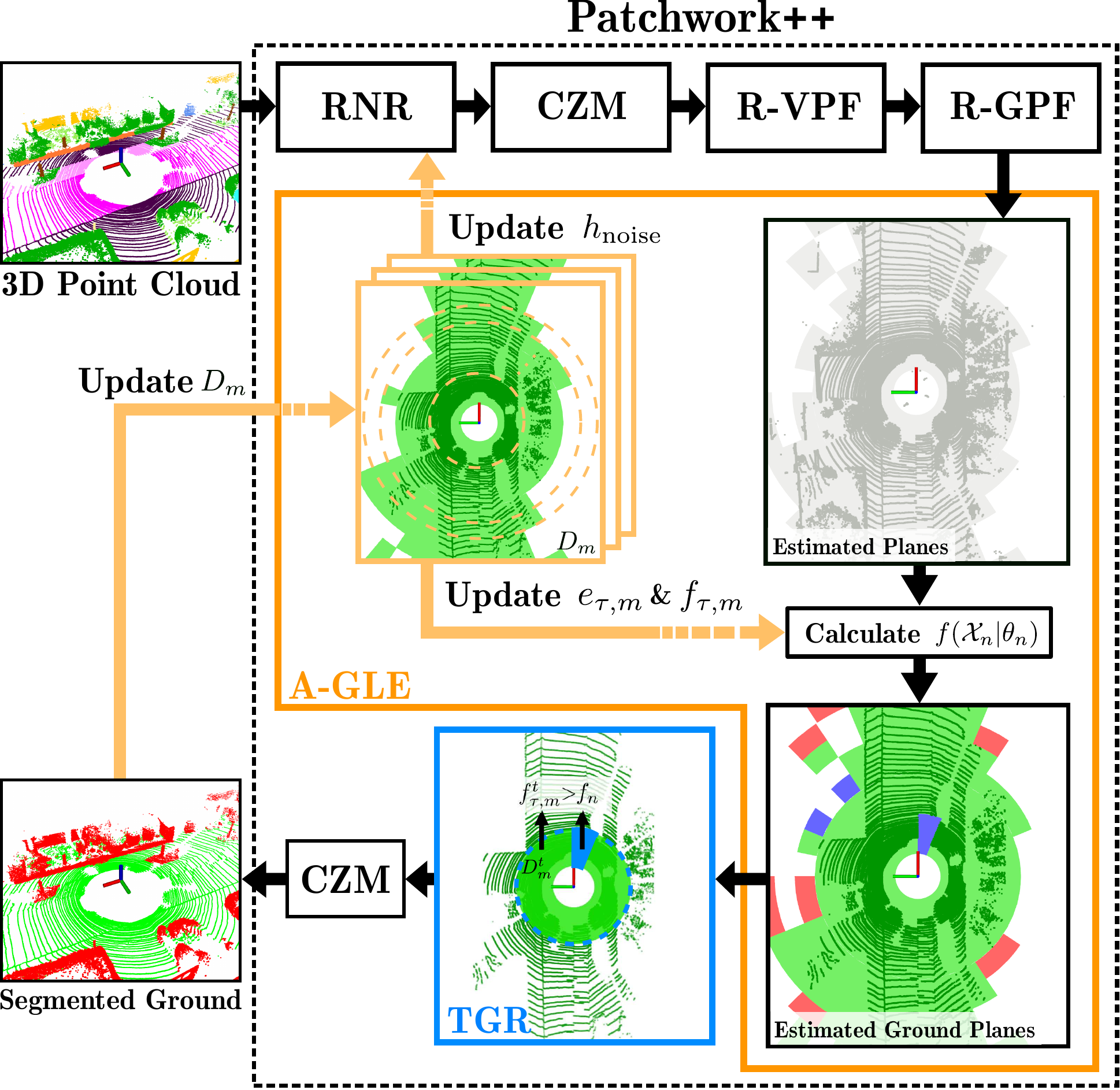}
	\vspace{-0.1cm}
	\caption{Visual description of A-GLE and TGR. A-GLE stores the variables of the estimated ground planes, i.e.,~$e_n$ and $f_n$, and updates its parameters based on the previous estimation results. TGR double-checks the under-segmented grounds based on the variables of the estimated ground planes at time $t$. }
	\label{fig:adaptive_gle}
\end{figure}

For better understanding, Fig.~\ref{fig:adaptive_gle} illustrates the flow in A-GLE. A-GLE stores the states from the previously estimated ground. Then, $h_{\text{noise}}$, $e_{\tau,m}$ and $f_{\tau,m}$ are updated for the next estimation based on the stored database.

\begin{figure*}[t!]
    \centering 
    \vspace{0.15cm}
	\begin{subfigure}[b]{0.1431\textwidth}
		\includegraphics[width=1.0\textwidth]{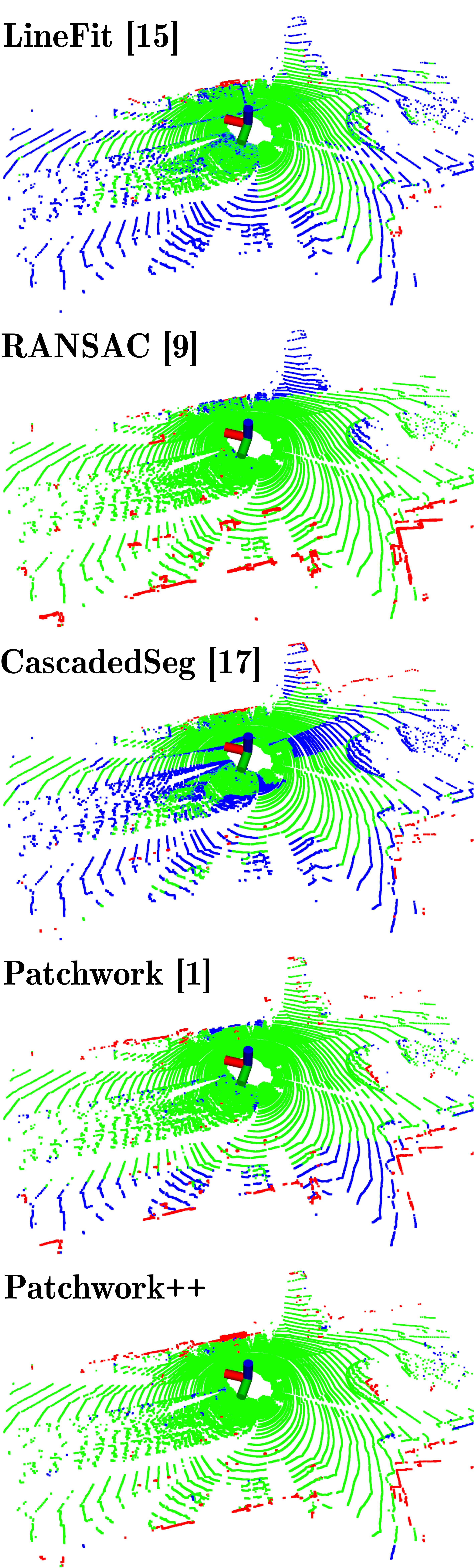}
		\caption{}
	\end{subfigure}
	\begin{subfigure}[b]{0.1431\textwidth}
		\includegraphics[width=1.0\textwidth]{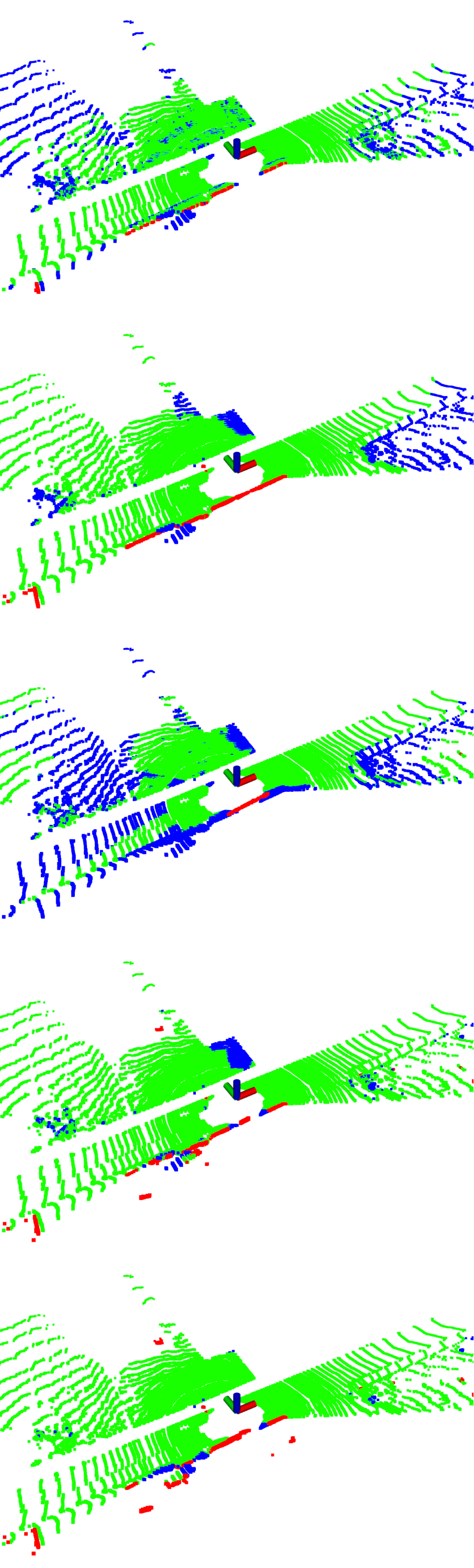}
		\caption{}
	\end{subfigure}
	\begin{subfigure}[b]{0.1431\textwidth}
		\includegraphics[width=1.0\textwidth]{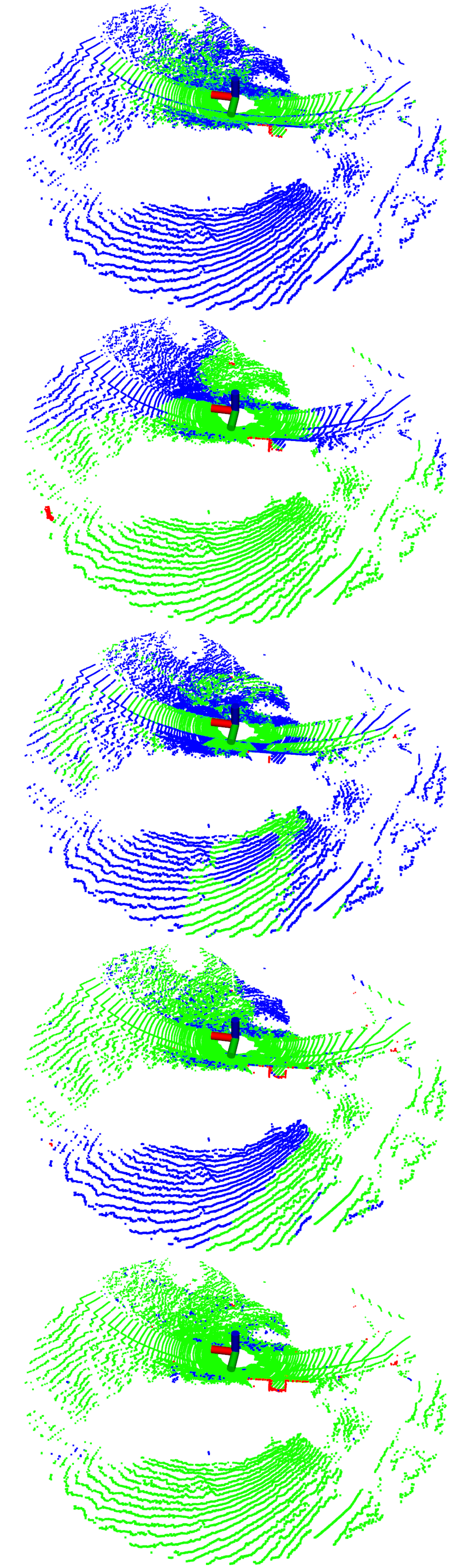}
		\caption{}
	\end{subfigure}
	\begin{subfigure}[b]{0.46\textwidth}
		\includegraphics[width=1.0\textwidth]{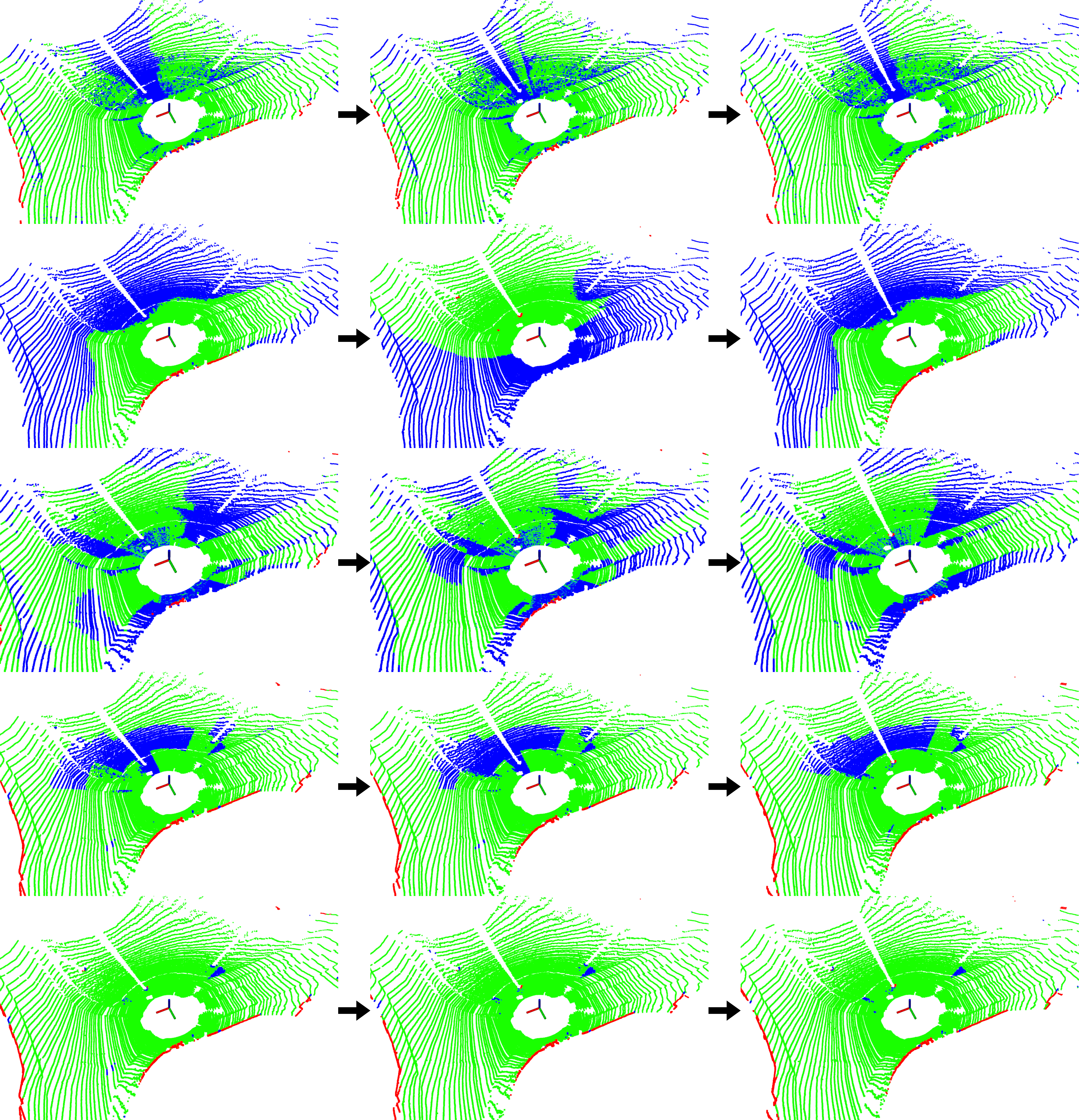}
		\caption{}
	\end{subfigure}
	\captionsetup{font=footnotesize}
	\vspace{-0.2cm}
	\caption{Qualitative comparison between the proposed method, Patchwork++, and the state-of-the-art methods at (a) Sequence \texttt{06} around frame 718, (b) Sequence \texttt{03} around frame 504, (c) Sequence \texttt{09} around frame 250, and (d) Sequence \texttt{02} around frame 3,300. Patchwork++ successfully addresses the partial under-segmentation problem. Green, red, and blue points represent $TP$, $FP$, and $FN$, respectively. The fewer the red and blue points, and the more the green points, the better~(best viewed in color).}
	\vspace{-0.6cm}
	\label{fig:sequential}
\end{figure*}








\subsection{TGR: Temporal Ground Revert}
\label{section:revert}

Through A-GLE, the estimation of typical grounds is successfully performed with a self-update of parameters. However, some unusual ground points within a bin do not satisfy the parameters' condition. For example, rough terrains with grass sometimes have higher $e_n$ and $f_n$ than the self-updated parameters. 
This is because A-GLE updates its parameters based on all values over time. Accordingly, this makes A-GLE play a role as a low-pass filter, which makes it difficult to estimate the bin as ground when $f_n$ becomes temporarily large.

To address this problem, TGR is proposed to revert the under-segmented ground planes into the segmented ground bins using the $D_m^t$, which are definite ground planes only at time $t$. In other words, TGR compares each under-segmented bin's $f_n$ with $f_{\tau,m}^t$, which can be calculated as follows:
\begin{equation}
    f_{\tau,m}^t = \text{mean} \big( F^t_{m} \big) + c_m\cdot \text{stdev} \big( F^t_{m} \big) ,
\end{equation} where $c_m$ denotes a constant gain of the standard deviation term for the $m$-th ring of the concentric zone model and $F_m^t$ is the set of $f_n$ in $D_m^t$. Therefore, the rejected ground points whose $f_n$ is larger than $f_{\tau,m}$ can be reverted if $f_n < f_{\tau,m}^t$ is satisfied (see Section~\rom{5}.\textit{D}).

\vspace{-0.15cm}

\section{Experimental Setup}

\subsection{Dataset and Error Metrics}


In our experiments, the SemanticKITTI dataset~\cite{behley2019semantickitti} was adopted to compare the ground segmentation performance of our proposed method to that of other ground segmentation algorithms. Based on the point-wise labels, the points labeled as \texttt{road}, \texttt{parking}, \texttt{sidewalk}, \texttt{other ground}, \texttt{lane marking}, and \texttt{terrain} are considered ground points. The other points are regarded as non-ground points. However, unlike our previous work~\cite{lim2021patchwork}, the points labeled as \texttt{vegetation} are not evaluated as ground nor non-ground exceptionally because it is \textcolor{000}{impractical} to regard the \texttt{vegetation} as a single ground or non-ground class. Note that this implies that points labeled as \texttt{vegetation} are only excluded in the evaluation step; the points are still included in the input point cloud. \textit{Precision}, \textit{Recall}, and \textit{$F_1$-score} are employed as error metrics as described in~\cite{lim2021patchwork}.

\subsection{Parameters of Patchwork++} 

We set initial $e_{\tau,m}$ and $f_{\tau,m}$ both $0$ for all $m$~($1\leq m \leq 4$). The other parameters are \textcolor{000}{identical to our} previous work~\textcolor{000}{\cite{lim2021patchwork}}. For RNR, we set $N_{\text{noise}}=20$ and $I_{\text{noise}}=0.2$. For R-VPF, we set $d_v=0.1$, $\theta_v=0.707$, and $K_v=3$. For A-GLE, we set $a_m=1$ for all $m$, $b_m=3$ for $m=1$, $b_m=2$ otherwise, and \textcolor{000}{$\delta=-0.5$}. For TGR, we set $c_m=1.5$ for all $m$. 


\section{Results and Discussion}

\begin{table}[b!]
    \centering
    \vspace{-0.5cm}
    \captionsetup{font=footnotesize}
    \caption{Performance comparison between our proposed method and other ground segmentation methods. Metrics are expressed as (mean)$\pm$(stdev).}
    \vspace{-0.2cm}
        \begin{tabular}{l|cc|c}
        \toprule \midrule
        {Algorithm} & {{Precision}~(\%)} & {{Recall}~(\%)} & { $F_1$~(\%) } \\
        \midrule
        LineFit~\cite{himmelsbach2010fast}   & \textbf{98.26}$\pm$\textbf{1.35} & 87.88$\pm$7.94 & 92.75 \\
        RANSAC~\cite{fischler1981random} & 89.87$\pm$14.16   & 93.97$\pm$13.16  & 91.83\\
        CascadedSeg~\cite{narksri2018slope}           & 95.25$\pm$7.88    & 74.53$\pm$10.78  & 83.59\\
        GPF~\cite{zermas2017fast}           & 95.78$\pm$3.76    & 83.89$\pm$22.42  & 89.14 \\
        R-GPF~\cite{lim2021erasor}                   & 74.68$\pm$15.7    & 98.15$\pm$\textbf{1.47}   & 84.52 \\
        Patchwork~\cite{lim2021patchwork}   & 94.23$\pm$3.96    & 97.62$\pm$3.42   & 95.88\\
        \midrule
        \textbf{Patchwork++ w/o TGR}                   & 94.98$\pm$3.44    & 97.64$\pm$3.58    & 96.28 \\ 
        \textbf{Patchwork++~(Ours)}         & 94.92$\pm$3.50    & \textbf{98.18}$\pm$2.41 & \textbf{96.51}\\ \midrule
        \bottomrule
        \end{tabular}
    \label{table:performance}
\end{table}

\subsection{Comparison with the State-of-the-Art Methods}

The overall performance of our proposed method and other ground segmentation methods are summarized in Table~\ref{table:performance} and Fig.~\ref{fig:sequential}. Our proposed algorithm exhibits the highest $F_1$-score compared with other ground segmentation methods, which include RANSAC~\cite{fischler1981random}, LineFit~\cite{himmelsbach2010fast}, GPF~\cite{zermas2017fast}, R-GPF~\cite{lim2021erasor}, and CascadedSeg~\cite{narksri2018slope}; our proposed method addresses the inherent problem of the previous noise filtering method~(Figs.~\ref{fig:sequential}(a) and (c)), the unsuitable definition of $f_n$ in the previous work~(Fig.~\ref{fig:sequential}(b)), and also the partial under-segmentation issue~(Figs.~\ref{fig:sequential}(c) and (d)). Consequently, Patchwork++ exhibits both higher precision and recall than Patchwork~\cite{lim2021patchwork}. Furthermore, the novel Patchwork++ presents a lower standard deviation of recall. These experimental evidences corroborate the position that our proposed method facilitates accurate and robust ground segmentation in complex urban environments.  




\subsection{Effect of R-VPF}


\begin{figure}[t!]
    \vspace{0.15cm}
	\begin{subfigure}[b]{0.15\textwidth}
		\includegraphics[width=1.0\textwidth]{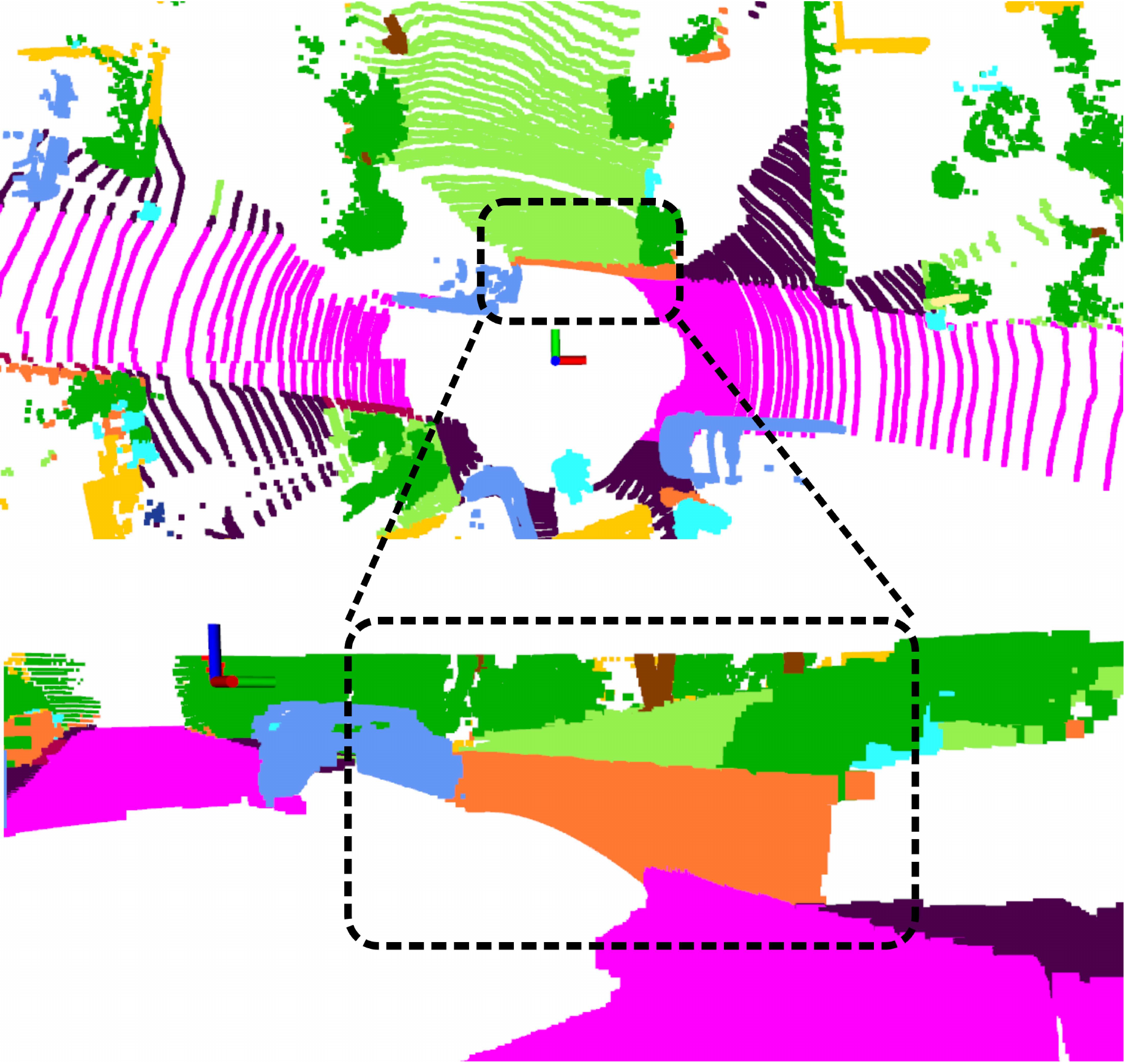}
		\caption{SemanticKITTI}
		\label{subfig:semantic}
	\end{subfigure}
	\centering
	\begin{subfigure}[b]{0.15\textwidth}
		\includegraphics[width=1.0\textwidth]{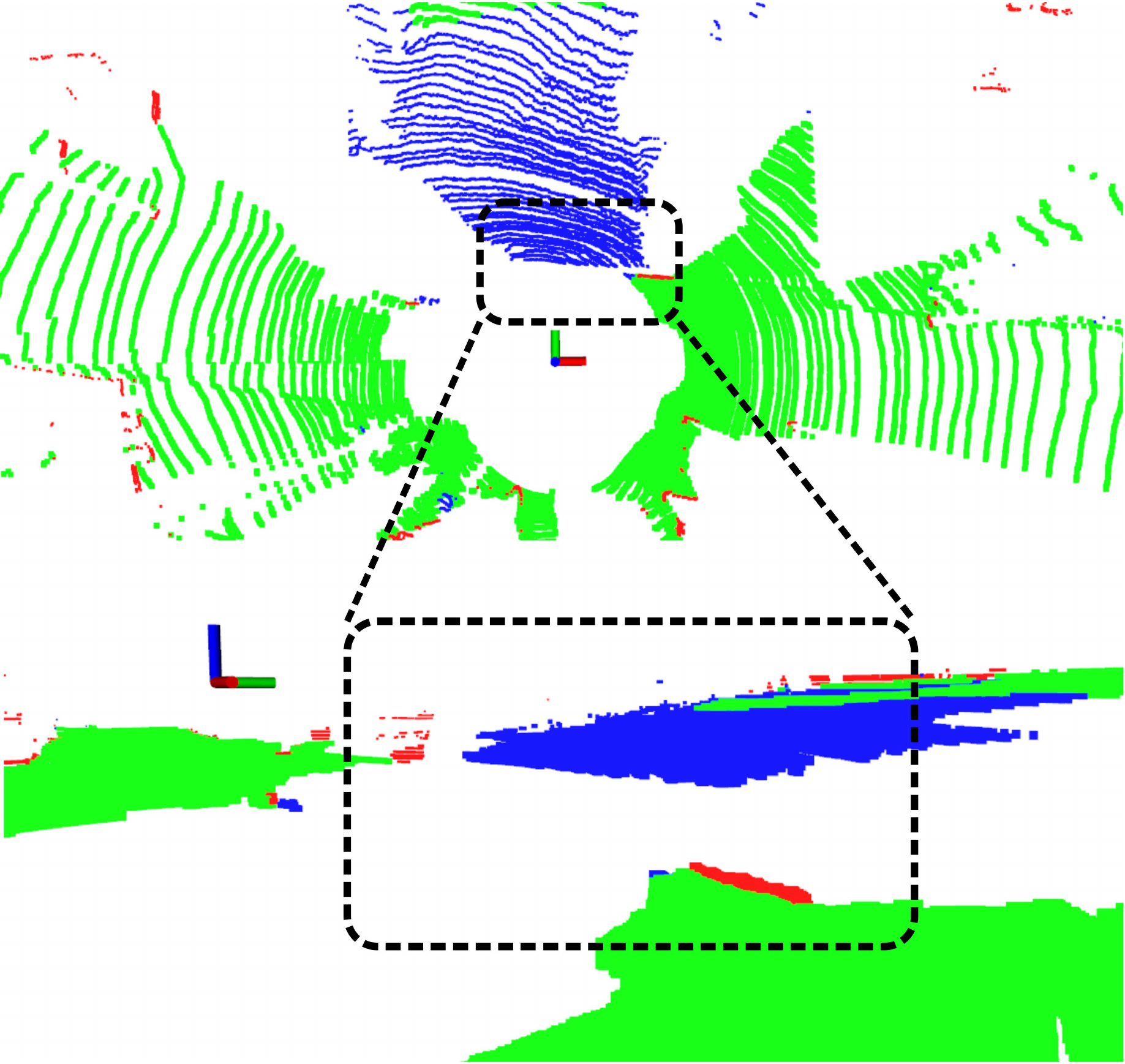}
		\caption{without R-VPF}
		\label{subfig:wo_r_vpf}
	\end{subfigure}
	\begin{subfigure}[b]{0.15\textwidth}
		\includegraphics[width=1.0\textwidth]{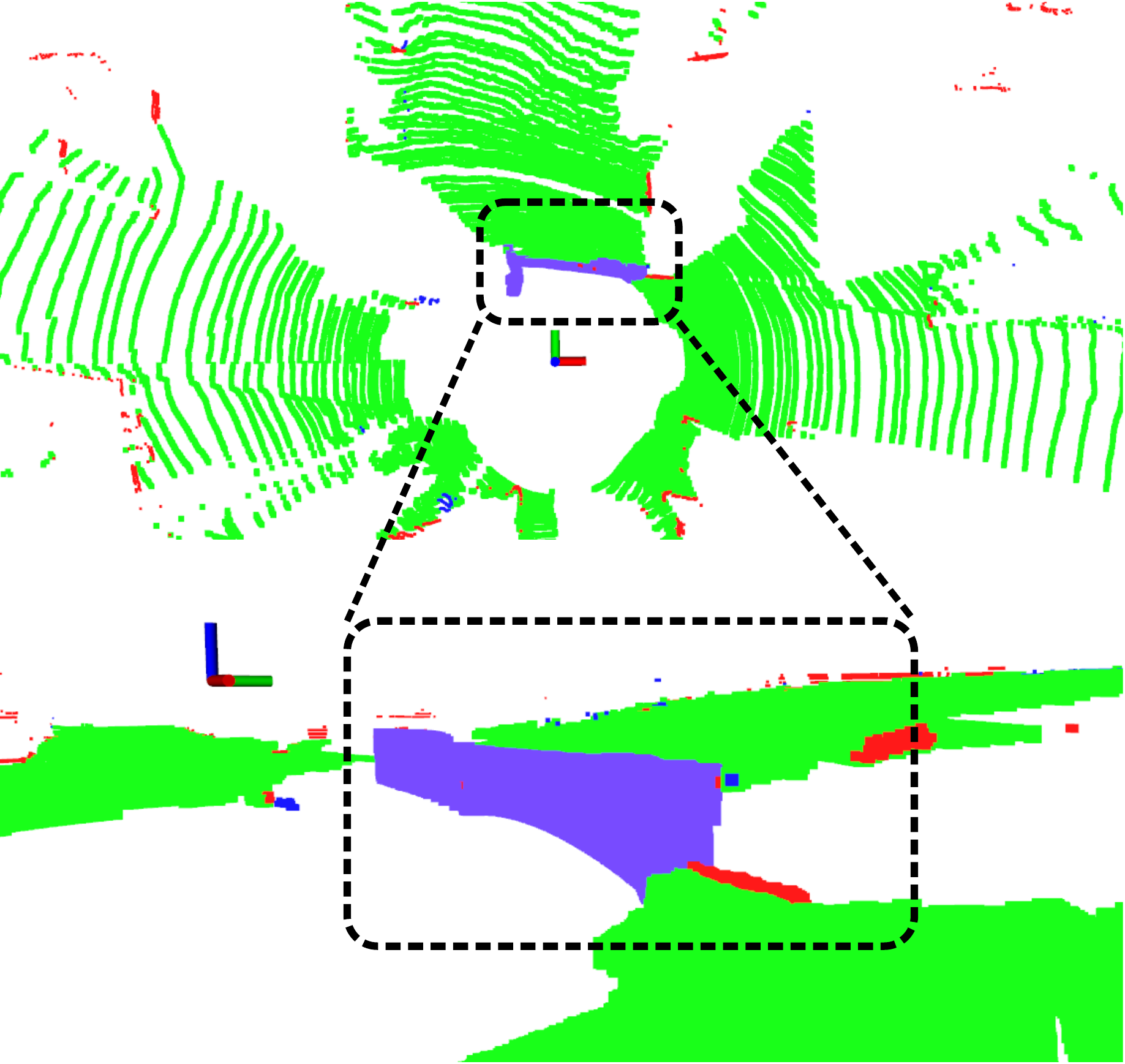}
		\caption{with R-VPF}
		\label{subfig:w_r_vpf}
	\end{subfigure}
    \captionsetup{font=footnotesize}
	\caption{(a) A visualized point cloud for Sequence \texttt{10} around frame 369 based on the labels in the SemanticKITTI dataset~\cite{behley2019semantickitti}. Note that the orange color denotes the points labeled as a fence. (b)-(c) Before and after the application of R-VPF. (b) The points from the fence impede R-GPF from estimating the correct ground plane, so the points from the terrain are under-segmented (blue). (c)~R-VPF \textcolor{000}{tackles} this problem by detecting vertical plane points in the preprocessing (purple), which is followed by successful ground segmentation by R-GPF even though ground points are located on the non-ground objects (best viewed in color).}
	\vspace{-0.6cm}
	\label{fig:r_vpf}
\end{figure}

\begin{figure}[b!]
    \vspace{-0.3cm}
    \captionsetup{font=footnotesize}
	\centering 
	\includegraphics[width=0.48\textwidth]{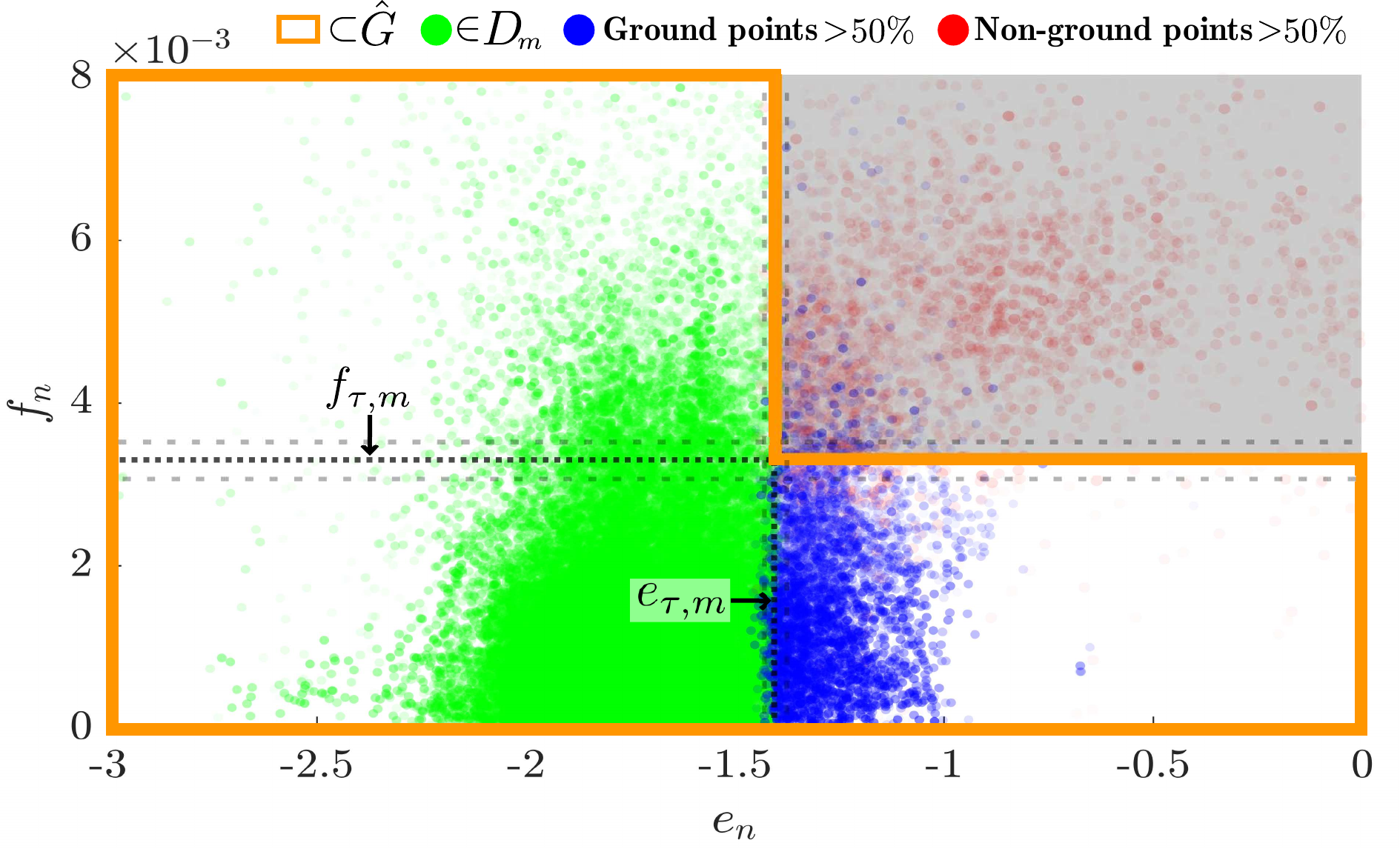}
	\caption{Results of A-GLE. The black-dotted vertical and horizontal lines represent the final elevation and flatness parameters, respectively. The gray-dashed lines represent the standard deviation of the self-updated parameters. It is verified that the self-updated flatness threshold $f_{\tau, m}$ is sufficiently valid as it classifies blue and red points appropriately. On the other hand, the validity of $e_{\tau,m}$ is ensured because approximately 95.8\% of the $D_m$ points are ground points. The opacity of plotted points represents the number of estimated plane points~(best viewed in color).}
	\label{fig:adjusted_flatness}
\end{figure}

In Fig.~\ref{fig:r_vpf}, it was demonstrated that R-VPF allows R-GPF to estimate correct ground planes even though the ground points are located on a low fence or flower bed. As mentioned earlier, R-GPF is strongly based on the premise that \textcolor{000}{the bottom points} are likely to be ground points; hence, R-GPF without R-VPF occasionally leads to lots of FN~(Fig.~\ref{fig:r_vpf}(b)). In contrast, R-GPF with R-VPF exhibits a successful ground segmentation by extracting points from the fence as non-ground points in advance~(Fig.~\ref{fig:r_vpf}(c)).



\subsection{Effect of A-GLE}

In Fig.~\ref{fig:adjusted_flatness}, the green points represent $D_m$ and approximately 95.8\% of them are true ground based on the SemanticKITTI dataset~\cite{behley2019semantickitti}. As the green points are determined by the self-updated $e_{\tau,m}$, this verifies that the resultant $e_{\tau,m}$ is sufficiently valid to estimate usual grounds and yields the appropriate $D_m$. Moreover, as 85.3\% of the true ground points are included in the green points, it is reasonable to update $f_{\tau, m}$ based on $D_m$, which classifies the blue and red points in Fig.~\ref{fig:adjusted_flatness}.

Similarly, the resultant $f_{\tau,m}$, which is drawn as a horizontal black-dotted line in Fig.~\ref{fig:adjusted_flatness}, appears to be suited to distinguish blue and red points, which can be assumed as ground and non-ground, respectively. For the remaining estimated planes, which do not belong to $D_m$, blue points represent the estimated planes where more than half are composed of the actual ground points, while red points represent the others. Hence, this verifies that the self-updated $f_{\tau,m}$ is sufficiently valid to distinguish ground planes from non-ground planes. 




\subsection{Effect of TGR}

\begin{figure}[t!]
    \vspace{0.15cm}
    \begin{subfigure}[b]{0.24\textwidth}
		\includegraphics[width=1.0\textwidth]{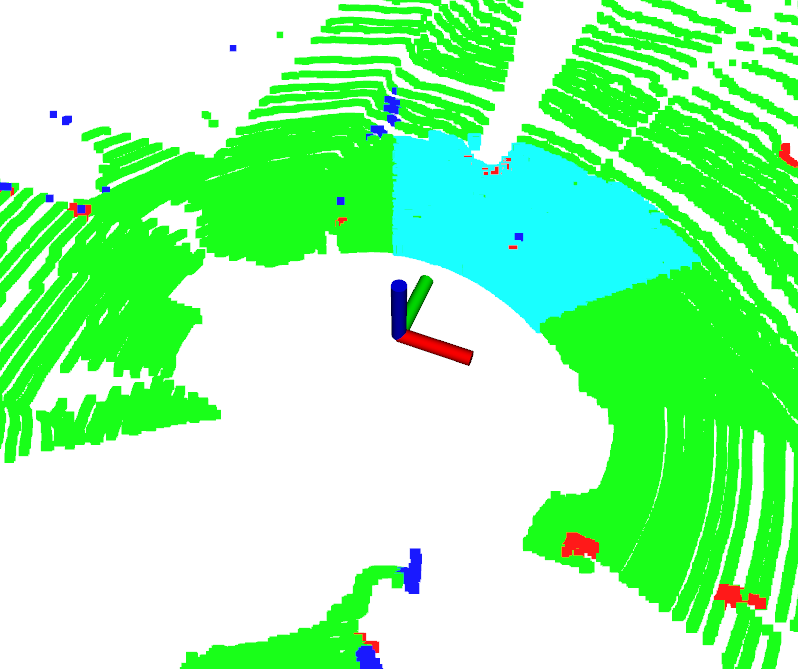}
	\end{subfigure}
	\begin{subfigure}[b]{0.24\textwidth}
		\includegraphics[width=1.0\textwidth]{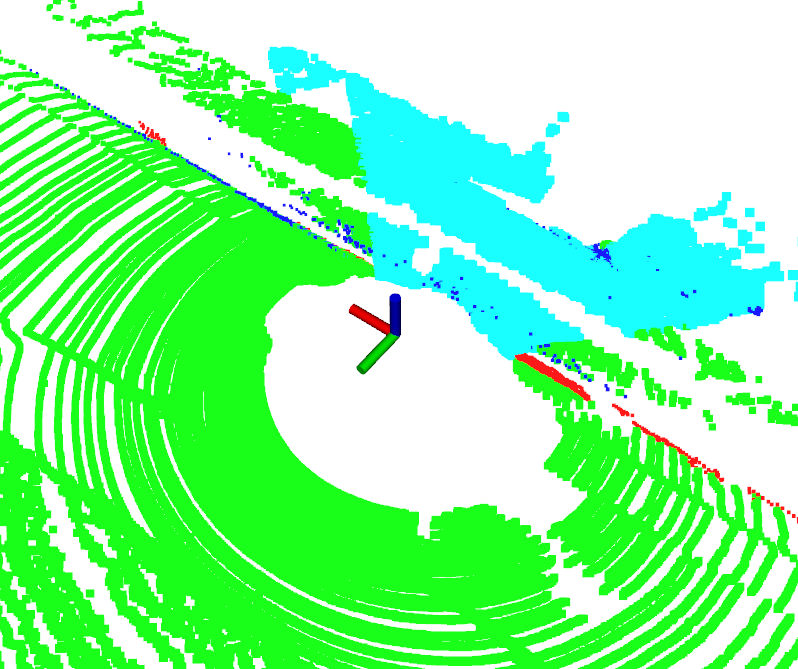}
	\end{subfigure}
    \captionsetup{font=footnotesize}
	\caption{(L-R) Examples of successful revert of ground points using TGR at Sequence \texttt{00} around frame 1,978 and Sequence \texttt{09} around frame 1,462. The cyan color denotes the reverted ground points, thus TGR significantly reduces false negatives~(best viewed in color).}
	\label{fig:tgr}
	\vspace{-0.7cm}
\end{figure}

Next, experimental evidences support that TGR resolves the partial under-segmentation issue, as shown in Fig.~\ref{fig:tgr}. Specifically, even though some bins have larger $f_n$ than $f_{\tau,m}$ in (\ref{eqn:self_update_thr_f}), these are double-checked based on the distribution characteristics of the currently estimated ground planes at time $t$. Therefore, TGR significantly reverts numerous FNs into TPs. Accordingly, TGR increases the recall with a negligible decrease in precision, as presented in Table~\ref{table:performance}. 

\subsection{Different Distributions of Self-Updated Parameters Depending on the Surroundings}

\begin{figure}[b!]
    \vspace{-0.1cm}
    \captionsetup{font=footnotesize}
	\centering 
	\begin{subfigure}[b]{0.23\textwidth}
		\includegraphics[width=1.0\textwidth]{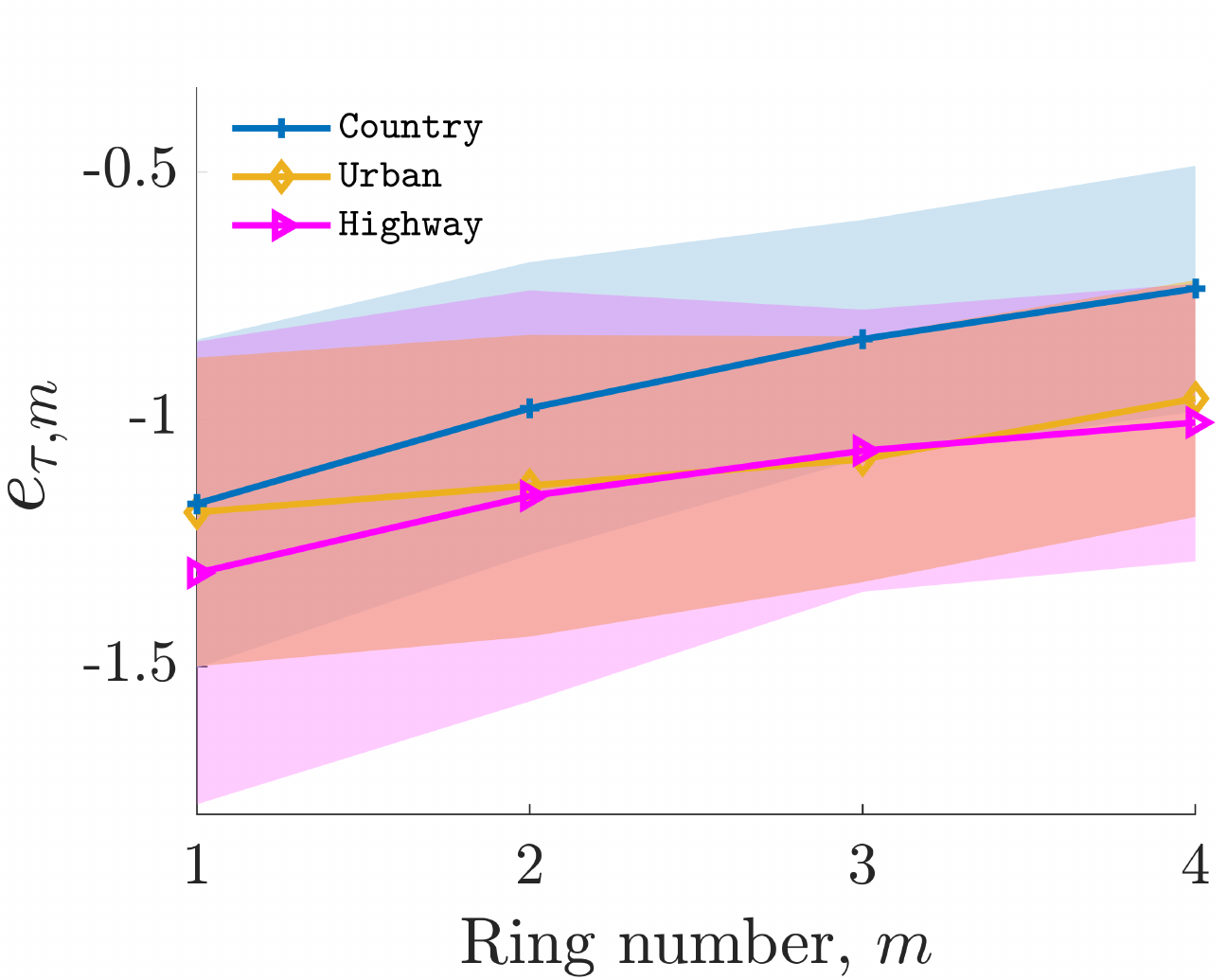}
		\caption{}
	\end{subfigure}
	\begin{subfigure}[b]{0.23\textwidth}
		\includegraphics[width=1.0\textwidth]{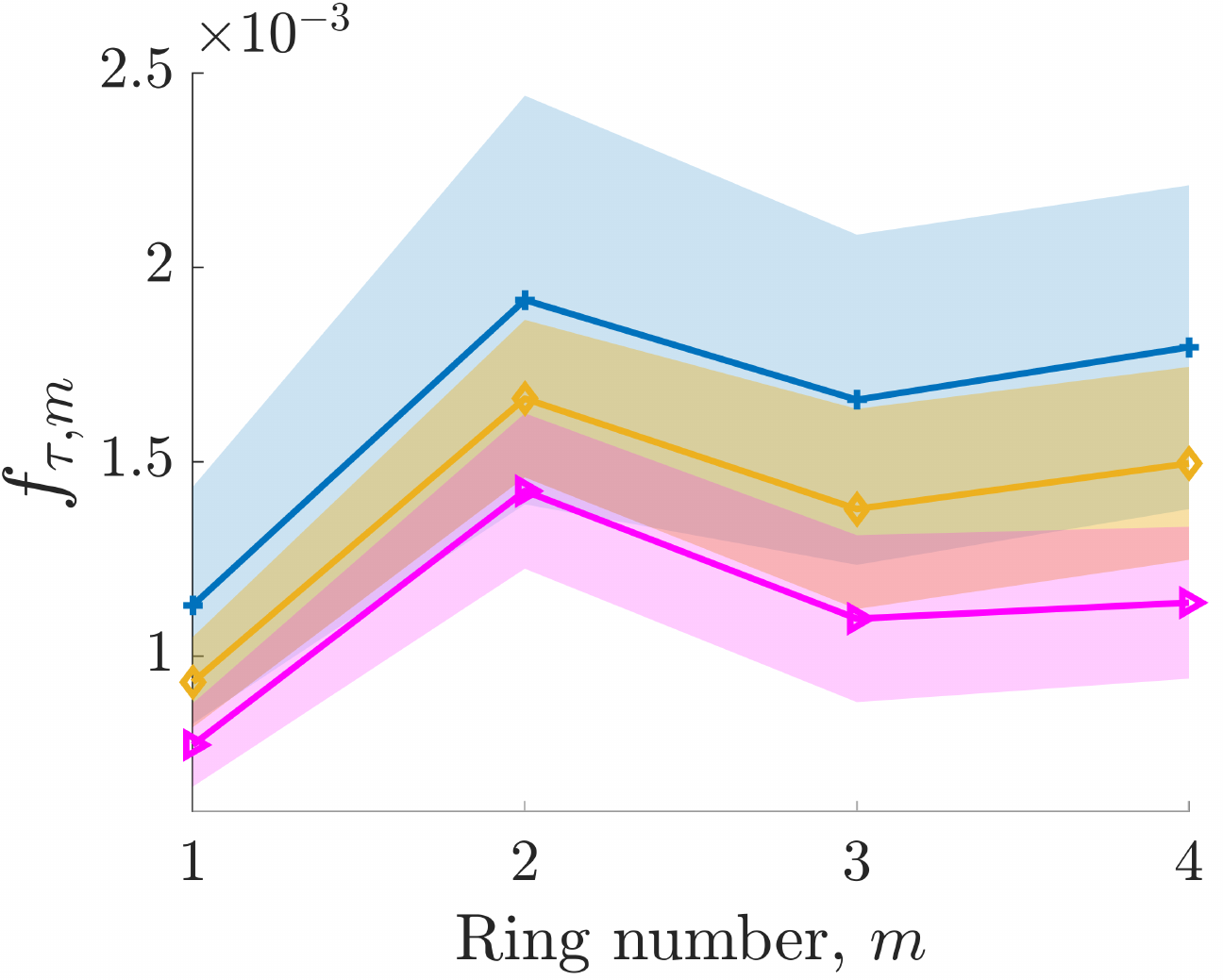}
		\caption{}
	\end{subfigure}
	\caption{Different distribution aspects of the self-updated (a) $e_{\tau,m}$ and (b) $f_{\tau, m}$ depending on the surroundings. Each colored area represents the standard deviation of self-updated parameters along the running time for the corresponding environment with the same color. \texttt{Country} includes Sequence \texttt{02}, \texttt{03}, \texttt{04}, \texttt{05}, \texttt{09}, and \texttt{10}; \texttt{Urban} includes Sequence \texttt{00}, \texttt{06}, \texttt{07}, and \texttt{08}; \texttt{Highway} denotes Sequence \texttt{01} in the SemanticKITTI dataset~\cite{behley2019semantickitti}. }
	\label{fig:adaptive_flatness_seq}
\end{figure}

Interestingly, it is determined that the self-updated parameters have different aspects depending on the surroundings, as presented in Fig.~\ref{fig:adaptive_flatness_seq}. For instance, in a highway scenario, the ground tends to be relatively flat compared with other scenarios, such that $e_{\tau,m}$ and $f_{\tau,m}$ are self-updated to small values. In contrast, country scenarios include relatively more steep roads, bumpy terrain, or sidewalks, such that $e_{\tau,m}$ and $f_{\tau,m}$ are set to higher values than the others. Therefore, it can be inferred that our A-GLE successfully updates the parameters depending on the surroundings, which reduces the effort required to update the parameter setting conducted by humans.



\subsection{Algorithm Speed} \label{sec:korea_univ}

Furthermore, our proposed method exhibits faster ground segmentation than Patchwork~\cite{lim2021patchwork}, which is noted in Table~\ref{table:speed}. This is an implementation-level improvement; originally, Patchwork employs global sorting based on the $z$ values before dividing a point cloud into CZM, which takes $O(N \log N)$ complexity, where $N$ denotes the size of the point cloud. In contrast, the sorting step in the currently proposed method is performed after the division of the point cloud into CZM; thus, its complexity can be approximated by $O(LM \log M)$, where $L$ and $M$ denote the number of bins and the number of points within each bin, respectively. Note that $N \gg L$ and $N \gg M$, and let us assume that $N=LM$ for simplicity. Then $N \log N$ can be expressed as $LM( \log L + \log M)$. Consequently, the bin-wise sorting can reduce the complexity by $LM \log L$.

\begin{table}[h]
    \centering
    \captionsetup{font=footnotesize}
    \caption{Speed comparison of ground segmentation methods on Sequence~\texttt{05} of the SemanticKITTI dataset, using Intel(R) Core(TM) i7-7700K CPU (unit: Hz).}
    \begin{tabular}{l|c|l|c}
    \toprule \midrule
    {Method} & {Speed} & {Method} & {Speed}\\
    \midrule
    LineFit~\cite{himmelsbach2010fast} & 58.96 & GPF~\cite{zermas2017fast} & 29.72 \\
    RANSAC~\cite{fischler1981random} & 15.43 & R-GPF~\cite{lim2021erasor} & 35.30 \\
    CascadedSeg~\cite{narksri2018slope} & 13.07 & Patchwork~\cite{lim2021patchwork} & 43.97 \\
    \midrule
    \textbf{Patchwork++~(Ours)} & 54.85 & \textbf{Patchwork++ w/o TGR} & \textbf{67.84} \\ \midrule
    \bottomrule
    \end{tabular}
    \label{table:speed}
\end{table}

\section{Conclusion}

In this study, a fast and robust ground segmentation method, \textit{Patchwork++}, has been proposed. The proposed method was verified to be faster and more robust than other state-of-the-art methods. Moreover, Patchwork++ reduces the number of parameters that require fine-tuning depending on the surroundings. Therefore, it has become easier to use ground segmentation in practice.

In the future, we plan to adopt Patchwork++ for various applications, including LiDAR odometry, object recognition, and static map building. Then, we will study the effect of ground segmentation on several applications.





\bibliographystyle{IEEEtran}
\bibliography{asap}

\begin{thebibliography}{10}
\providecommand{\url}[1]{#1}
\csname url@rmstyle\endcsname
\providecommand{\newblock}{\relax}
\providecommand{\bibinfo}[2]{#2}
\providecommand\BIBentrySTDinterwordspacing{\spaceskip=0pt\relax}
\providecommand\BIBentryALTinterwordstretchfactor{4}
\providecommand\BIBentryALTinterwordspacing{\spaceskip=\fontdimen2\font plus
\BIBentryALTinterwordstretchfactor\fontdimen3\font minus
  \fontdimen4\font\relax}
\providecommand\BIBforeignlanguage[2]{{%
\expandafter\ifx\csname l@#1\endcsname\relax
\typeout{** WARNING: IEEEtran.bst: No hyphenation pattern has been}%
\typeout{** loaded for the language `#1'. Using the pattern for}%
\typeout{** the default language instead.}%
\else
\language=\csname l@#1\endcsname
\fi
#2}}

\bibitem{lim2021patchwork}
H.~Lim, M.~Oh, and H.~Myung, ``Patchwork: Concentric zone-based region-wise
  ground segmentation with ground likelihood estimation using a \text{3D LiDAR}
  sensor,'' \emph{IEEE Robot. Autom. Lett.}, vol.~6, no.~4, pp. 6458--6465,
  2021.

\bibitem{lim2021erasor}
H.~Lim, S.~Hwang, and H.~Myung, ``\text{ERASOR}: Egocentric ratio of pseudo
  occupancy-based dynamic object removal for static \text{3D} point cloud map
  building,'' \emph{IEEE Robot. Autom. Lett.}, vol.~6, no.~2, pp. 2272--2279,
  2021.

\bibitem{li2021lidar}
Z.~Li, F.~Wang, and N.~Wang, ``\text{LiDAR R-CNN}: An efficient and universal
  \text{3D} object detector,'' in \emph{Proc. IEEE/CVF Conf. Comput. Vis.
  Pattern Recognit.}, 2021, pp. 7546--7555.

\bibitem{qi2017pointnet}
C.~R. Qi, H.~Su, K.~Mo, and L.~J. Guibas, ``\text{PointNet}: Deep learning on
  point sets for \text{3D} classification and segmentation,'' in \emph{Proc.
  IEEE Comput. Soc. Conf. Comput. Vis. Pattern Recognit.}, 2017, pp. 652--660.

\bibitem{sung2021if}
C.~Sung, S.~Jeon, H.~Lim, and H.~Myung, ``{What if there was no revisit?
  Large-scale graph-based SLAM with traffic sign detection in an HD map using
  LiDAR inertial odometry},'' \emph{Intell. Service Robot.}, pp. 1--10, 2021.

\bibitem{koide2021voxelized}
K.~Koide, M.~Yokozuka, S.~Oishi, and A.~Banno, ``Voxelized \text{GICP} for fast
  and accurate \text{3D} point cloud registration,'' in \emph{Proc. IEEE Int.
  Conf. Robot. Autom.}, 2021, pp. 11\,054--11\,059.

\bibitem{suger2015traversability}
B.~Suger, B.~Steder, and W.~Burgard, ``Traversability analysis for mobile
  robots in outdoor environments: A semi-supervised learning approach based on
  \text{3D LiDAR} data,'' in \emph{Proc. IEEE Int. Conf. Robot. Autom.}, 2015,
  pp. 3941--3946.

\bibitem{yan2017online}
Z.~Yan, T.~Duckett, and N.~Bellotto, ``Online learning for human classification
  in \text{3D LiDAR}-based tracking,'' in \emph{Proc. IEEE/RSJ Int. Conf.
  Intell. Robots Syst.}, 2017, pp. 864--871.

\bibitem{fischler1981random}
M.~A. Fischler and R.~C. Bolles, ``Random sample consensus: A paradigm for
  model fitting with applications to image analysis and automated
  cartography,'' \emph{Commun. ACM}, vol.~24, no.~6, pp. 381--395, 1981.

\bibitem{xu2021rpvnet}
J.~Xu, R.~Zhang, J.~Dou, Y.~Zhu, J.~Sun, and S.~Pu, ``\text{RPVnet}: A deep and
  efficient range-point-voxel fusion network for \text{LiDAR} point cloud
  segmentation,'' in \emph{Proc. IEEE/CVF Int. Conf. Comput. Vis.}, 2021, pp.
  16\,024--16\,033.

\bibitem{paigwar2020gndnet}
A.~Paigwar, {\"O}.~Erkent, D.~Sierra-Gonzalez, and C.~Laugier, ``\text{GndNet}:
  Fast ground plane estimation and point cloud segmentation for autonomous
  vehicles,'' in \emph{Proc. IEEE/RSJ Int. Conf. Intell. Robots Syst.}, 2020,
  pp. 2150--2156.

\bibitem{behley2019semantickitti}
J.~Behley, M.~Garbade, A.~Milioto, J.~Quenzel, S.~Behnke, C.~Stachniss, and
  J.~Gall, ``\text{SemanticKITTI}: A dataset for semantic scene understanding
  of \text{LiDAR} sequences,'' in \emph{Proc. IEEE/CVF Int. Conf. Comput.
  Vis.}, 2019, pp. 9297--9307.

\bibitem{shen2021fast}
Z.~Shen, H.~Liang, L.~Lin, Z.~Wang, W.~Huang, and J.~Yu, ``Fast ground
  segmentation for \text{3D LiDAR} point cloud based on
  \text{Jump-Convolution-Process},'' \emph{Remote Sensing}, vol.~13, no.~16, p.
  3239, 2021.

\bibitem{moosmann2009segmentation}
F.~Moosmann, O.~Pink, and C.~Stiller, ``Segmentation of \text{3D LiDAR} data in
  non-flat urban environments using a local convexity criterion,'' in
  \emph{Proc. IEEE Intell. Veh. Symp.}, 2009, pp. 215--220.

\bibitem{himmelsbach2010fast}
M.~Himmelsbach, F.~V. Hundelshausen, and H.-J. Wuensche, ``Fast segmentation of
  \text{3D} point clouds for ground vehicles,'' in \emph{Proc. IEEE Intell.
  Veh. Symp.}, 2010, pp. 560--565.

\bibitem{zermas2017fast}
D.~Zermas, I.~Izzat, and N.~Papanikolopoulos, ``Fast segmentation of \text{3D}
  point clouds: A paradigm on \text{LiDAR} data for autonomous vehicle
  applications,'' in \emph{Proc. IEEE Int. Conf. Robot. Autom.}, 2017, pp.
  5067--5073.

\bibitem{narksri2018slope}
P.~Narksri, E.~Takeuchi, Y.~Ninomiya, Y.~Morales, N.~Akai, and N.~Kawaguchi,
  ``A slope-robust cascaded ground segmentation in \text{3D} point cloud for
  autonomous vehicles,'' in \emph{Proc. IEEE Int. Conf. Intell. Transp. Syst.},
  2018, pp. 497--504.

\bibitem{jimenez2021ground}
V.~Jim{\'e}nez, J.~Godoy, A.~Artu{\~n}edo, and J.~Villagra, ``Ground
  segmentation algorithm for sloped terrain and sparse \text{LiDAR} point
  cloud,'' \emph{IEEE Access}, vol.~9, pp. 132\,914--132\,927, 2021.

\bibitem{li2020insclustering}
Y.~Li, C.~Le~Bihan, T.~Pourtau, and T.~Ristorcelli, ``Insclustering: Instantly
  clustering \text{LiDAR} range measures for autonomous vehicle,'' in
  \emph{Proc. IEEE Int. Conf. Intell. Transp. Syst.}, 2020, pp. 1--6.

\bibitem{sualeh2019dynamic}
M.~Sualeh and G.-W. Kim, ``Dynamic multi-\text{LiDAR} based multiple object
  detection and tracking,'' \emph{Sensors}, vol.~19, no.~6, p. 1474, 2019.

\bibitem{asvadi20163d}
A.~Asvadi, C.~Premebida, P.~Peixoto, and U.~Nunes, ``\text{3D LiDAR}-based
  static and moving obstacle detection in driving environments: An approach
  based on voxels and multi-region ground planes,'' \emph{Robot. Auton. Syst.},
  vol.~83, pp. 299--311, 2016.

\bibitem{shan2018lego}
T.~Shan and B.~Englot, ``{LeGO-LOAM: Lightweight and ground-optimized
  \text{LiDAR} odometry and mapping on variable terrain},'' in \emph{Proc.
  IEEE/RSJ Int. Conf. Intell. Robots Syst.}, 2018, pp. 4758--4765.

\bibitem{pan2021mulls}
Y.~Pan, P.~Xiao, Y.~He, Z.~Shao, and Z.~Li, ``\text{MULLS}: Versatile
  \text{LiDAR SLAM} via multi-metric linear least square,'' in \emph{Proc. IEEE
  Int. Conf. Robot. Autom.}, 2021, pp. 11\,633--11\,640.

\bibitem{lim2022pago}
D.-U. Seo, H.~Lim, S.~Lee, and H.~Myung, ``{PaGO-LOAM: Robust ground-optimized
  LiDAR odometry},'' in \emph{Proc. Int. Conf. Ubiquit. Robot.}, 2021,
  {Submitted}.

\bibitem{zhao2020mapping}
X.~Zhao, Z.~Yang, and S.~Schwertfeger, ``Mapping with reflection -- detection
  and utilization of reflection in \text{3D LiDAR} scans,'' in \emph{Proc. IEEE
  Int. Symp. Saf. Secur. Rescue Robot.}, 2020, pp. 27--33.

\bibitem{weinmann2015semantic}
M.~Weinmann, B.~Jutzi, S.~Hinz, and C.~Mallet, ``Semantic point cloud
  interpretation based on optimal neighborhoods, relevant features and
  efficient classifiers,'' \emph{ISPRS J. Photogramm. Remote Sens.}, vol. 105,
  pp. 286--304, 2015.

\end{thebibliography}

\end{document}